\documentclass[lettersize,journal]{IEEEtran}
\usepackage{amsmath,amsfonts}
\usepackage{algorithmic}
\usepackage{algorithm}
\usepackage{array}
\usepackage[caption=false,font=normalsize,labelfont=sf,textfont=sf]{subfig}
\usepackage{textcomp}
\usepackage{stfloats}
\usepackage{url}
\usepackage{verbatim}
\usepackage{graphicx}
\usepackage{cite}
\usepackage[breaklinks=true,bookmarks=true]{hyperref}
\usepackage{CJKutf8}
\usepackage{amsthm,amsmath,amssymb,lipsum}
\usepackage{mathrsfs}
\usepackage{amssymb}

\usepackage{threeparttable}
\usepackage{tabularx}
\usepackage{color}
\usepackage{booktabs}
\usepackage{verbatim}
\usepackage{amsopn}
\usepackage{amsmath}
\usepackage{multirow}
\usepackage{colortbl}
\usepackage{makecell} 

\usepackage{colortbl,xcolor}

\usepackage{pdfpages}

\usepackage{xspace}
\makeatletter
\DeclareRobustCommand\onedot{\futurelet\@let@token\@onedot}
\def\@onedot{\ifx\@let@token.\else.\null\fi\xspace}
\def\eg{\emph{e.g}\onedot} 
\def\ie{\emph{i.e}\onedot} 
 
\def\etc{\emph{etc}\onedot} 
\def\myvs{\emph{vs}\onedot}
 
\def\etal{\emph{et al}\onedot}
\makeatother

\usepackage{overpic}

\makeatletter
\newcommand{\thickhline}{%
 \noalign {\ifnum 0=`}\fi \hrule height 1pt
 \futurelet \reserved@a \@xhline
}
\makeatother
\newcommand{\pub}[1]{\color{gray}{\tiny{[{#1}]}}}

\usepackage{amsmath,amsfonts,bm}

\def\eqref#1{equation~\ref{#1}}

\def\1{\bm{1}}

\DeclareMathAlphabet{\mathsfit}{\encodingdefault}{\sfdefault}{m}{sl}
\SetMathAlphabet{\mathsfit}{bold}{\encodingdefault}{\sfdefault}{bx}{n}

\usepackage{pifont}

\begin{document}

\title{Task-Generalized Adaptive Cross-Domain Learning for Multimodal Image Fusion}

\author{Mengyu Wang, Zhenyu Liu, Kun Li, Yu Wang, Yuwei Wang, Yanyan Wei, Fei Wang$^\ast$  

\thanks{This work was supported by the National Natural Science Foundation of China (62465014, U24A20331) and Natural Science Foundation of Jiangxi Province of China (20232BCJ23096, 20232BAB212016, 20224BAB202006, 20203BBE53038, GJJ2401004).

\textit{$^*$Corresponding author: Fei Wang}}

\thanks{
M. Wang is with the Key Laboratory of Opto-Electronic Information Science and Technology of Jiangxi Province, Nanchang Hangkong University, Nanchang, Jiangxi, 330063, China, and also with the Key Laboratory of Nondestructive Test (Ministry of Education), Nanchang Hangkang University, Nanchang, Jiangxi, 330063, China. (e-mail: mengyu@nchu.edu.cn).}
\thanks{
Z. Liu is with the Key Laboratory of Opto-Electronic Information Science and Technology of Jiangxi Province, Nanchang Hangkong University, Nanchang, Jiangxi, 330063, China. (e-mail: 2408085408337@stu.nchu.edu.cn).}
\thanks{
K. Li is with the ReLER, CCAI, Zhejiang University, Hangzhou, 310027, China (e-mail: kunli.hfut@gmail.com)
}
\thanks{
Y. Wang and Y. Wang are with the College of Engineering, Anhui Agricultural University, Hefei, 230036, China. (e-mail: wyw@ahau.edu.cn). 
}
\thanks{
Y. Wei and F. Wang are with the School of Computer Science and Information Engineering, Hefei University of Technology, Hefei, 230601, China. (e-mails: weiyy@hfut.edu.cn; jiafei127@gmail.com). 
}
}

\markboth{Journal of \LaTeX\ Class Files,~Vol.~14, No.~8, August~2021}%
{Shell \MakeLowercase{\textit{et al.}}: A Sample Article Using IEEEtran.cls for IEEE Journals}

\maketitle

\begin{abstract}
Multimodal Image Fusion (MMIF) aims to integrate complementary information from different imaging modalities to overcome the limitations of individual sensors. It enhances image quality and facilitates downstream applications such as remote sensing, medical diagnostics, and robotics. Despite significant advancements, current MMIF methods still face challenges such as modality misalignment, high-frequency detail destruction, and task-specific limitations. To address these challenges, we propose AdaSFFuse, a novel framework for task-generalized MMIF through adaptive cross-domain co-fusion learning. AdaSFFuse introduces two key innovations: the Adaptive Approximate Wavelet Transform (AdaWAT) for frequency decoupling, and the Spatial-Frequency Mamba Blocks for efficient multimodal fusion. AdaWAT adaptively separates the high- and low-frequency components of multimodal images from different scenes, enabling fine-grained extraction and alignment of distinct frequency characteristics for each modality. The Spatial-Frequency Mamba Blocks facilitate cross-domain fusion in both spatial and frequency domains, enhancing this process. These blocks dynamically adjust through learnable mappings to ensure robust fusion across diverse modalities.
By combining these components, AdaSFFuse improves the alignment and integration of multimodal features, reduces frequency loss, and preserves critical details. Extensive experiments on four MMIF tasks—Infrared-Visible Image Fusion (IVF), Multi-Focus Image Fusion (MFF), Multi-Exposure Image Fusion (MEF), and Medical Image Fusion (MIF)—demonstrate AdaSFFuse's superior fusion performance, ensuring both low computational cost and a compact network, offering a strong balance between performance and efficiency.
The code will be publicly available at \url{https://github.com/Zhen-yu-Liu/AdaSFFuse}.
\end{abstract}


\section{Introduction}\label{sec:intro}
\IEEEPARstart{M}{ultimodal} Image Fusion (MMIF) integrates complementary information from multiple sources to overcome the limitations of individual imaging devices~\cite{zhao2023cddfuse,Zhao_2024_CVPR,ma2022swinfusion}. It is widely applied in fields such as remote sensing, medical imaging, and robotics, with common tasks including Infrared-Visible Image Fusion (IVF)~\cite{jian2021infrared,li2021different,li2023lrrnet}, Multi-Exposure Image Fusion (MEF)~\cite{wu2022dmef,xu2020mef}, Multi-Focus Image Fusion (MFF)~\cite{zhao2018multi,nie2022mlnet}, and Medical Image Fusion (MIF)~\cite{tang2022matr,wen2023msgfusion}. Infrared images capture thermal features under low-light conditions but are limited by low resolution and lack color, whereas visible images offer higher resolution but less prominent target features. Combining these modalities compensates for their respective shortcomings, enhancing the visibility of key targets. MEF preserves fine details in high dynamic range scenarios, while MEF addresses blur caused by depth-of-field limitations. Similarly, MIF integrates modalities, such as CT and MRI, leveraging complementary information to improve diagnostic accuracy.

Traditional MMIF methods~\cite{liu2015general} predominantly rely on clustering or multi-scale transforms. However, their performance is often limited by complex scenes, noise, and manual feature extraction, leading to information loss or distortion. 
Recently, learning-based methods~\cite{ma2020ddcgan,ma2022swinfusion,liu2022target}, particularly unsupervised multi-scale fusion networks, have gained significant attention. 
Despite these advancements, conventional temporal visual encoding modules struggle to capture high-frequency details and handle modality differences effectively. This challenge is especially evident in MFF, where the intricate textures of foreground and background pose difficulties for architectures like Transformer~\cite{wang2024low,li2023vigt}. These limitations highlight the difficulty of fitting spatial domain information from different modalities into a universal MMIF network.

\begin{figure*}[t]
\centering
\includegraphics[width=1.0\linewidth]{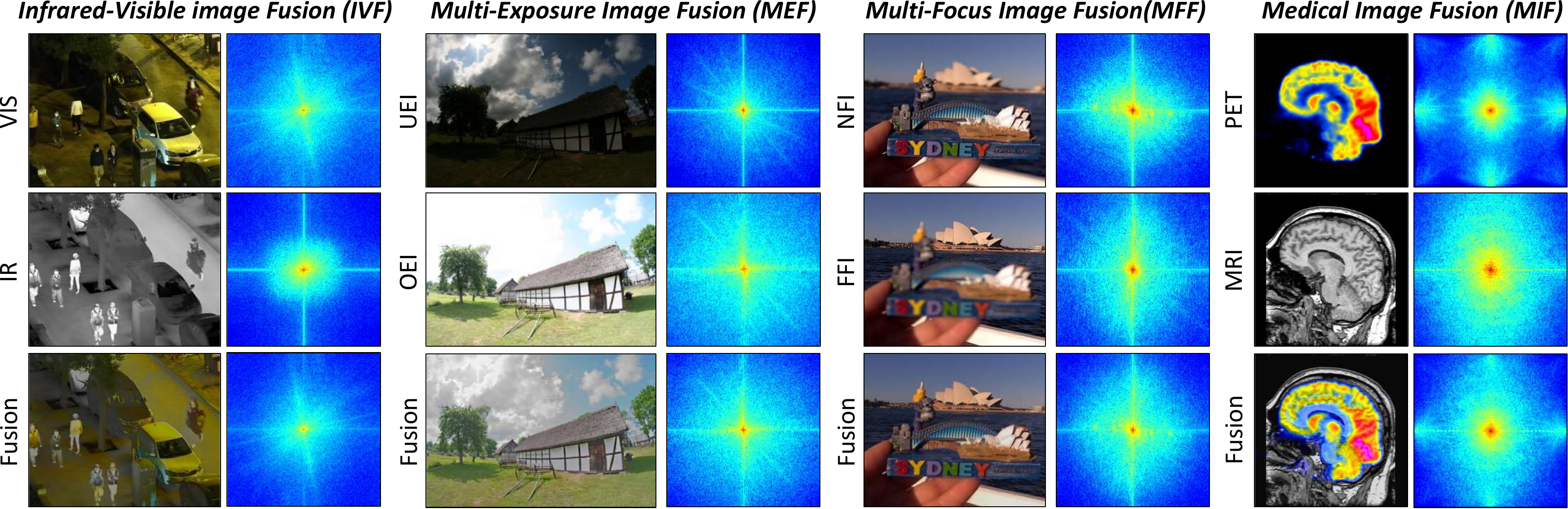}
\caption{Frequency domain visualization of four typical MMIF tasks: Infrared-Visible Fusion (IVF), Multi-Exposure Fusion (MEF), Multi-Focus Fusion (MFF), and Medical Image Fusion (MIF). The frequency spectra of different modality images and their fusion results exhibit a clear pattern of frequency component aggregation, with the fused image spectrum capturing the diverse frequency characteristics of the original modalities.}
\label{Fig:fft}
\end{figure*}

Frequency information can be viewed as a combination of high- and low-frequency components in natural images~\cite{chen2019drop}. A simple and straightforward approach is to use frequency-domain transformations in place of spatial-domain ones, which not only offers more algorithmic options but also directs attention to different frequency components.
Consequently, we conducted frequency analysis on four MMIF tasks, as shown in Figure~\ref{Fig:fft},
Typically, high-frequency components in RGB images are primarily manifested in regions with significant brightness gradients (\eg, edges, textures).
In contrast, infrared (IR) imaging is highly sensitive to thermal radiation, emphasizing regions with pronounced temperature differences. IR images inherently contain richer high-frequency components than nighttime RGB images due to their unique ability to capture temperature-driven structural details, which are less affected by ambient lighting or surface color variations.
Through spatial-domain and Fourier spectral analysis, we found that MMIF primarily targets the dominant features of each modality to achieve complementarity information. In other words, each modality exhibits distinct frequency characteristics in the spectrum.

As a frequency prior, the Discrete Fourier Transform (DFT)~\cite{solachidis2007watermarking,narwaria2012fourier,du2025improving} provides frequency-domain information for reconstructing desired high- and low-frequency components.
Cui~\etal~\cite{cui2023image} decoupled images via frequency selection and demonstrated the significant advantage of frequency-domain information in distinguishing subtle modality differences. Shen~\etal~\cite{shen2024spatial} incorporated fast Fourier transform (FFT) in Transformers for frequency decoupling learning. In comparison, Discrete Wavelet Transform (DWT)~\cite{fernandes2003new} is more suitable for handling images with sharp signal changes. Studies~\cite{WANG2021114574,9005243,9018389} have shown that wavelet decomposition and similar frequency decoupling methods perform efficiently in MMIF tasks. Additionally, task-related frequency decoupling learning has gained more attention.
SFNet~\cite{cui2023selective} combines dynamic convolution with a multi-branch structure to decouple high-frequency and low-frequency. Wang~\etal~\cite{wang2024frequency} used dilated convolutions to approximate frequency decoupling, capturing task-adaptive frequency components.
This facilitates precise structural preservation and targeted feature enhancement, enabling complementary cross-modal integration in line with multi-domain learning goals.

Based on the above analysis, we investigate the challenges of generic MMIF and propose our improvement as follows:

\begin{itemize}
\item\textbf{Explorations of Generic Frequency Decoupling.}
Most frequency decoupling models, such as Laplace pyramids, Fourier filters, and wavelet transforms, rely on hand-crafted techniques that are non-learnable and often task-agnostic. These limitations may hinder the extraction of valuable information from modality-specific images.
An alternative approach is to leverage the frequency characteristics to capture the desired frequency features. For instance, average pooling and dilated convolutions with large dilation rates act as low-pass filters, while max pooling and small kernel convolutions function as high-pass filters. 
However, these methods fail to integrate high- and low-frequency components effectively. 
In this work, we propose combining group convolutions and dilated convolutions within a 2D-DWT framework to match the spatial distribution of the data adaptively. This approach transforms the data into the wavelet domain, generating sub-signal features with different frequency bands and orientations.
Similarly, a corresponding inverse 2D-DWT design is then employed to recover frequency information, thereby optimizing image reconstruction accuracy. 

\item\textbf{Explorations of Adaptive Domain Fusion.} 
Inspired by recent MMIF methods~\cite{zhao2023cddfuse,ma2022swinfusion,li2024mambadfuse}, we adopt a shallow-deep fusion approach, performing frequency-band fusion between the high- and low-frequency components of different modalities.
This method enhances the interaction and frequency alignment between modality images, addressing the frequency loss and collapse issues observed in traditional fusion models.
However, existing methods typically focus only on high-dimensional features in the spatial domain, lacking effective adaptation to frequency features.
Directly using these features often results in a decline in frequency information and fusion performance.
Transformer-based MMIF models~\cite{ma2022swinfusion,sun2024image} strengthen feature interaction through stacking, while Mamba’s state space model (SSM) improves global information modeling with a selection mechanism and hardware-aware design, maintaining linear complexity. Recent studies~\cite{li2024mambadfuse,peng2024fusionmamba} show that Mamba outperforms traditional methods in both performance and efficiency for MMIF.
Building on this advantage, we propose integrating an adaptive spatial-frequency mapping module into the 2D-State Space Duality (SSD) module of the Mamba2 architecture, further enhancing the adaptation and fusion capabilities of different frequency features.
\end{itemize}

Based on the above discussions, we propose AdaSFFuse, as shown in Figure~\ref{Fig:overall}, a framework consisting of two key components: the Adaptive Approximate Wavelet Transform (AdaWAT) module and the Spatial-Frequency Mamba Blocks.
AdaSFFuse tackles the challenges in generic multimodal image fusion by overcoming performance discrepancies across different task modalities. 
Specifically, for any pair of multimodal image feature embeddings, AdaSFFuse first performs frequency decoupling through AdaWAT, which decomposes the image into multi-band sub-signals that represent low-frequency global information and high-frequency detail features, enabling fine-grained separation of modality-specific features.
Then, frequency-segmented fusion is applied to the multi-band sub-signals of each modality to strengthen frequency consistency and promote inter-modality collaboration.
Finally, the fused result is restored to the original feature space using the Adaptive Approximate Inverse Wavelet Transform (AdaIWAT).
Building upon this, the Spatial-Frequency Mamba Blocks leverage spatial awareness and frequency filtering capabilities to support the Adaptive Domain State Space Duality (AdaD-SSD), which solves cross-domain linear mapping and effectively promotes fusion.
It dynamically adapts to different frequencies and modalities for efficient fusion, operating on both shallow features following frequency-segmented fusion and deep features after Ada-IWAT.
Overall, the main contributions are summarized below:
\begin{itemize} 
\item We propose a novel paradigm for task-generalized MMIF, named AdaSFFuse, which integrates adaptive multi-domain co-fusion to bridge the gap of existing methods in multimodal image co-modeling.
\item We design an adaptive approximate wavelet transform for decoupling and reconstructing multi-band sub-signals, enabling fine-grained extraction of cross-modal information.
\item We introduce a spatial-frequency Mamba block, which empowers the advanced Mamba architecture with learnable spatial awareness and frequency filtering capabilities to facilitate efficient and synergistic multi-domain fusion.
\item Extensive qualitative and quantitative experiments on four MMIF tasks (IVF, MFF, MEF, and MIF) demonstrate superior performance and robustness of our AdaSFFuse.
\end{itemize}

\section{Related Work}

The advances of deep learning have significantly improved image fusion~\cite{zhao2023cddfuse,Zhao_2024_CVPR,liu2024task,liu2024searching}, surpassing traditional methods by leveraging robust feature extraction and generalization.
Early works~\cite{liu2017multi,ram2017deepfuse,zhang2020ifcnn,wang2023interactively,liu2022attention,liu2021learning} laid the groundwork for deep learning-based image fusion.
Subsequent studies~\cite{9879642,wang2023interactively} advanced the integration of infrared and visible modalities using adversarial learning and multi-scene validation, paving the way for learnable multimodal fusion.
Practical concerns were also addressed~\cite{wang2024improving,liu2024infrared,lei2025mlfuse,wang2025robust}, such as alignment errors, through progressive dense alignment and adaptive fusion strategies. Moreover, \cite{Liu_2023} employed coupled contrastive learning and multi-level integration to enhance fusion quality and downstream task performance.
While DRDEC~\cite{sun2024deep} enhanced feature representation with residual architectures.
The field is further advanced with GANs-based methods~\cite{ma2020ddcgan}. DDFM~\cite{zhao2023ddfm} innovated with dual-branch dense feature mining, and HitFusion~\cite{chen2024hitfusion} incorporated hierarchical feature transformation. MUFusion~\cite{cheng2023mufusion} introduced a unified multi-task framework for comprehensive feature learning.
Additionally, the advent of attention mechanisms and Transformers~\cite{ma2022swinfusion,li2022cgtf} marked a paradigm shift, improving generative quality. 
Specifically, EMMA~\cite{Zhao_2024_CVPR} incorporated modality-aware extraction, while ITFuse~\cite{tang2024itfuse} proposed instance transformation. CDDFuse~\cite{zhao2023cddfuse} enabled cross-domain feature extraction, SwinFusion~\cite{ma2022swinfusion} modeled long-range dependencies with Swin Transformer blocks~\cite{wang2025exploiting}, and SHIP~\cite{zheng2024probing} adopts a high-level interactive collaboration mechanism to effectively complement information.
Recently, MambaDFuse~\cite{li2024mambadfuse} achieved SOTA performance, offering improved efficiency in capturing long-range dependencies for fusion.


\subsection{Adaptive Frequency Learning}
A main research focus has been effectively integrating frequency information into neural network architectures to learn potential patterns efficiently~\cite{2024freqfusion,wang2024frequency,shen2024spatial,huang2022winnet}.
Traditional frequency domain methods, such as Fast Fourier Transform (FFT)~\cite{shen2024spatial} and Wavelet Transform (WAT)~\cite{huang2022winnet}, typically process frequency information in a static manner.
Although these methods perform well in certain tasks, they fail to adjust to image content and task-specific requirements dynamically, limiting their applicability in complex scenarios~\cite{dong2023head,yun2023spanet}.
In contrast, recent advances~\cite{2024freqfusion,dong2023head,wang2024frequency} emphasize adaptive frequency mechanisms within a learning-based framework, enabling dynamic adjustment and decoupling of frequency information. 
For example, \cite{shen2024spatial,2024freqfusion,huang2022winnet} integrated FFT or WAT descriptors into learnable modules for multi-domain perception. Similarly, \cite{wang2024frequency,liang2023omni,chen2019drop} explored convolutional approximations of Gaussian or wavelet kernels for frequency decomposition, adapting to different tasks. \cite{shen2024spatial,si2022inception} investigated the frequency characteristics of representative modules, such as low- and high-pass capabilities. 
In this work, we extend WAT operators and learning-based modules for both feature perception and frequency decoupling.

\subsection{Visual Mamba Model}
State Space Models (SSMs)~\cite{mamba} provide a mathematical framework for describing dynamic systems, capturing input-output relationships through hidden states.
Compared to traditional Transformers~\cite{vaswani2017attention,wang2024eulermormer,li2025repetitive}, SSMs are particularly effective in modeling continuous long sequences and have been widely applied in various tasks. For example, Vamba~\cite{zhu2024vision} integrated bidirectional SSMs into a general-purpose visual backbone for positional encoding and visual representation compression. Dang~\etal~\cite{dang2024log} enhanced Mamba's local perception capability for medical image segmentation by incorporating both local and global information. Li~\etal~\cite{li2024mambadfuse} proposed MambaDFuse, a two-stage model that effectively integrates complementary information from different modalities using a dual-layer feature extractor and a two-phase feature fusion module. The recent State Space Duality (SSD) architecture~\cite{mamba2} significantly improves Mamba by offering substantial advantages in computational complexity and efficiency. 
However, task-generalized MMIF, particularly in modeling cross-task modality fusion, remains underexplored and presents significant challenges.

\section{Methodology}
\begin{figure*}[t!]
\centering
\includegraphics[width=1.0\linewidth]{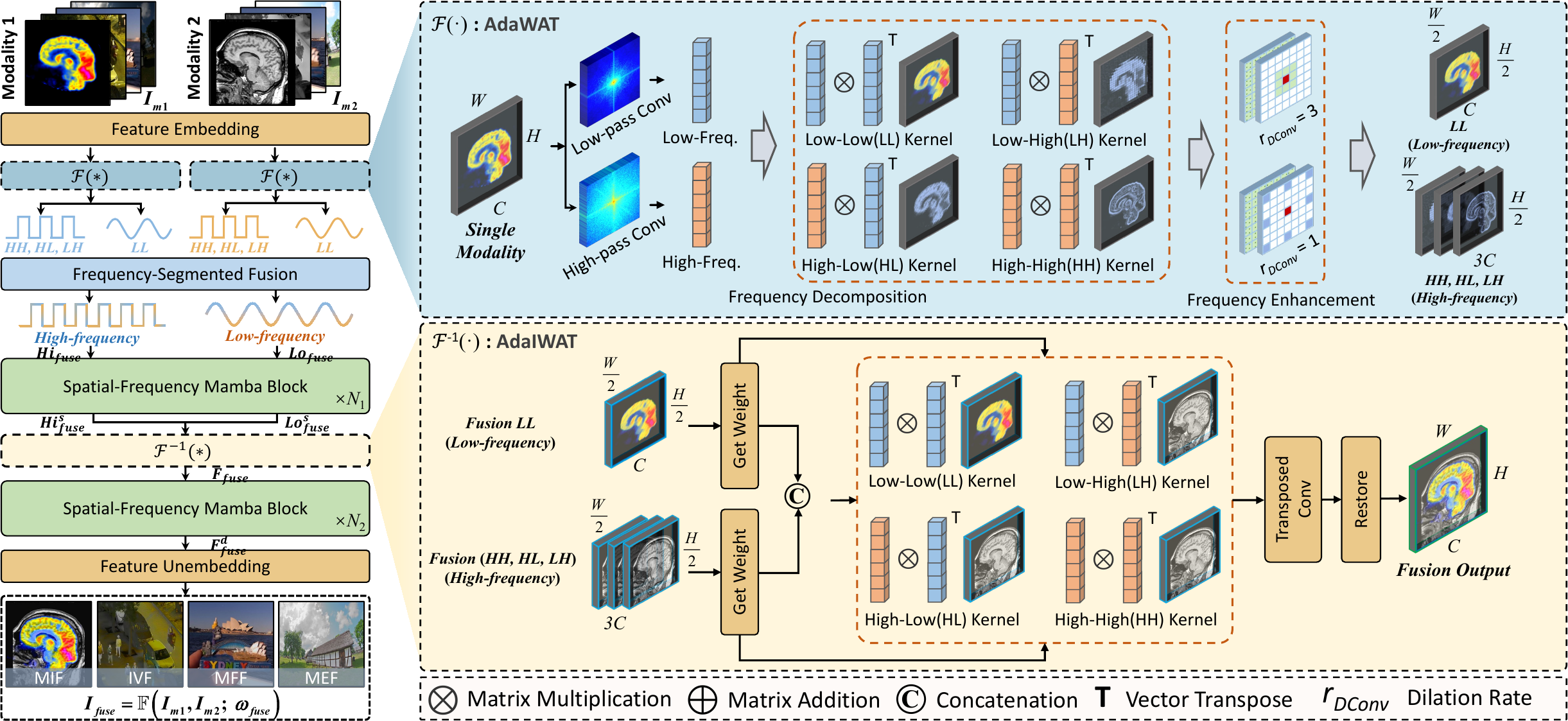}
\caption{Overall architecture of the proposed AdaSFFuse network is designed for task-generalized multimodal image fusion (MMIF). It leverages the Adaptive Approximate Wavelet Transform (AdaWAT) to achieve efficient frequency decoupling, adaptable to various fusion scenarios. This allows for the separation and processing of low- and high-frequency features from different image modalities.
In addition, AdaSFFuse integrates the Spatial-Frequency Mamba Block, a fusion module specifically tailored for MMIF, incorporating the core Adaptive Domain SSD (AdaD-SSD) component, which enables domain adaptation feature fusion by modeling both spatial and frequency information, leading to high-quality fused images with enhanced alignment and complementary details.
} 
\label{Fig:overall}
\end{figure*}

\subsection{Overall Framework}
The overall pipeline diagram of the proposed AdaSFFuse network, $\mathbb{F}(\cdot)$, is shown in Figure~\ref{Fig:overall}. For the task-generalized MMIF task, given two aligned source images from different modalities (assumed to be grayscale images), $\mathbf{I}_{m1} \in \mathbb{R}^{H \times W \times  1}$ and $\mathbf{I}_{m2} \in \mathbb{R}^{H \times W \times  1}$, the function $\mathbb{F}(\cdot)$ is designed to learn an optimal set of parameters $\omega_{fuse}$ that yield the fused image $\mathbf{I}_{fuse}$=$\mathbb{F}( \mathbf{I}_{m1},\mathbf{I}_{m2};\omega_{fuse})$.

Our proposed AdaSFFuse first utilizes AdaWAT to decouple valuable high- and low-frequency sub-signal features $\{\mathcal{F}_{LH_{{\{m1,m2}\}}},\mathcal{F}_{HL_{{\{m1,m2}\}}},\mathcal{F}_{HH_{{\{m1,m2}\}}}\}$$\!\in \mathbb{R}^{\frac{H}{4}\times \frac{W}{4} \times C}$ and $\mathcal{F}_{LL_{{\{m1,m2}\}}}$$\!\in\mathbb{R}^{\frac{H}{4}\times \frac{W}{4} \times  C}$ between initial feature embeddings $\mathbf{F}_{m1}\!\in\!\mathbb{R}^{\frac{H}{2}\times \frac{W}{2} \times  C}$ and $\mathbf{F}_{m2}\!\in\!\mathbb{R}^{\frac{H}{2}\times \frac{W}{2} \times  C}$ in a way that is suitable for frequency decomposition in a generic MMIF task.
Then, the high- and low-frequency features of different modalities are fused by specific frequency-segmented fusion to from $\textbf{Hi}_{fuse}$=$\sum_{n=1}^2\{\mathcal{F}_{LH_{m_n}},\mathcal{F}_{HL_{m_n}},\mathcal{F}_{HH_{m_n}}\}\in\mathbb{R}^{\frac{H}{4}\times \frac{W}{4} \times  3C}$ and $\textbf{Lo}_{fuse}$=$\sum_{n=1}^2\mathcal{F}_{LL_{{\{m1,m2}\}}}\in\mathbb{R}^{\frac{H}{4}\times \frac{W}{4} \times C}$, which are then fed into the Spatial-Frequency Mamba module for shallow fusion of the adaptive frequency features, yielding $\textbf{Hi}^{s}_{fuse}\in\mathbb{R}^{\frac{H}{4}\times \frac{W}{4} \times  3C}$ and $\textbf{Lo}^{s}_{fuse}\in\mathbb{R}^{\frac{H}{4}\times \frac{W}{4} \times  C}$.

Finally, AdaIWAT, in conjunction with the Spatial-Frequency Mamba, fully exploits the complementary information in both the spatial and frequency domains to guide the deep fusion $\textbf{F}^{d}_{fuse}\in\mathbb{R}^{\frac{H}{4}\times \frac{W}{4} \times  C}$, ultimately reconstructing the fused image $\textbf{I}_{fuse}\in\mathbb{R}^{H \times W \times  1}$.
We will describe two key components: the Adaptive Approximate Wavelet Transform (AdaWAT) and the Spatial-Frequency Mamba.

\begin{figure}[t!]
\centering
\includegraphics[width=1.0\linewidth]{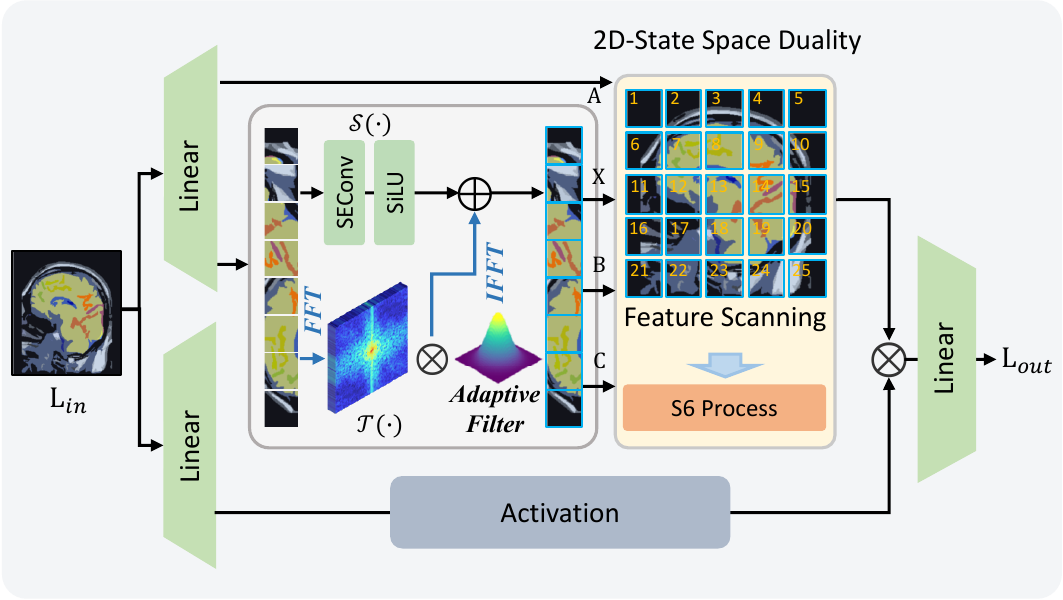}
\caption{Details of Adaptive Domain State Space Dualit (AdaD-SSD), including two key improvements: 1) a spatial-aware and frequency filtering module, and 2) an enhanced 2D-SSD architecture, improving the network’s ability to process and fuse diverse frequency features.}
\label{fig:mamba}
\end{figure}

\subsection{Adaptive Approximate Wavelet Transform}\label{adawat}
To address the limitations of the traditional wavelet transform in spatial-frequency resolution and task adaptation, we propose an Adaptive Approximate Wavelet Transform (AdaWAT) designed to adaptively fit the spatial-frequency domain transform for multimodal image inputs and auto-adjust the wavelet bases, enabling more efficient frequency-domain decoupling and feature extraction to optimize the subsequent MMIF task.

In AdaWAT, we redefine the process of constructing the wavelet transform to accommodate a wider range of signal processing needs.
The case of a 1D signal is given first.
Given an input discrete signal $\mathit{t}$, a scale factor $\mathit{j}$, a time factor $\mathit{k}$, and a scaling function $\mathit{\varphi _{j} }$, we define the wavelet function $\psi_{j,k}(t)=2^{j/2}\psi(2^{j}t-k)$ and the scale function $\phi_{j,k}(t) = 2^{j/2}\phi(2^{j}t - k)$, and the decomposition of the signal at scale $\mathit{j _{0} }$, as:
\begin{equation}
x(t) = \sum_{j > j_0} \sum_k w_{j,k} \psi_{j,k}(t) + \sum_k v_{j_0,k} \phi_{j_0,k}(t),
\end{equation}
where $w_{j,k}$=$\left \langle x(t),\psi_{j,k}(t) \right \rangle$ denotes the detail coefficients and $v_{j_{0},k}$=$\left \langle x(t),\phi_{j_{0},k}(t)\right \rangle$ denotes the approximation coefficients.
Next, we introduced the adaptive analysis vectors $\vec{u}_0[k]$=$2^{1/2}\phi(2t - k)$ and $\vec{u}_1[k] = 2^{1/2} \psi(2t - k)$, corresponding to the low and high-frequency filters, respectively.
The decomposition of the signal in the wavelet basis can be viewed as a recursive convolution with a particular filter:
\begin{equation}
\begin{aligned}
v_{j+1,p} &= \sum_k \vec{u}_0[k - 2p] a_{j,k},\\ 
w_{j+1,p} &= \sum_k \vec{u}_1[k - 2p] a_{j,k}.
\end{aligned}
\end{equation}

For a 2D digital image with any of the input image features $\textbf{F}_{m1}\in\mathbb{R}^{\frac{H}{2}\times \frac{W}{2}\times C}$, we extend this decomposition to denote:
\begin{equation}	  
\begin{aligned}  
\textbf{K}_{\mathcal{F}_{LL}} & = \vec{u}_0 \times \vec{u}_0^T; &\textbf{K}_{\mathcal{F}_{LH}} & = \vec{u}_0 \times \vec{u}_1^T; \\  
\textbf{K}_{\mathcal{F}_{HL}} & = \vec{u}_1 \times \vec{u}_0^T; &\textbf{K}_{\mathcal{F}_{HH}} & = \vec{u}_1 \times \vec{u}_1^T;
\end{aligned}  
\end{equation}

where $\textbf{K}_{\mathcal{F}_{LL}}\in\mathbb{R}^{ \frac{H}{4}\times  \frac{W}{4}\times C}$, $\textbf{K}_{\mathcal{F}_{LH}}\in\mathbb{R}^{ \frac{H}{4}\times  \frac{W}{4}\times C}$, $\textbf{K}_{\mathcal{F}_{HL}}\in\mathbb{R}^{ \frac{H}{4}\times  \frac{W}{4}\times C}$ and 
$\textbf{K}_{\mathcal{F}_{HH}}\in\mathbb{R}^{ \frac{H}{4}\times  \frac{W}{4}\times C}$ denote the convolution operators for low frequency, horizontal high frequency, vertical high frequency, and diagonal high frequency, respectively, implemented through grouped convolutions~\cite{krizhevsky2012imagenet}.
The wavelet kernel is formed by concatenating these convolution operators.

In addition, to enhance frequency-specific information, we apply dilated convolutions with different dilation rates to each frequency component~\cite{wang2024frequency,chen2024frequency,zhao2025temporal}. The low-frequency component is enhanced with a 3$\times$3 dilated convolution (dilation rate $r_{DConv}$=$3$) for low-pass filtering, improving smoothness and expressiveness. The high-frequency component is enhanced with a 3$\times$3 dilated convolution (dilation rate $r_{DConv}$=$1$) for high-pass filtering, emphasizing details and resolution. This results in decoupled high- and low-frequency features, 
$\{\mathcal{F}_{LH_{m1}},\mathcal{F}_{HL_{m1}},\mathcal{F}_{HH_{m1}}\}$ and $\mathcal{F}_{LL_{m1}}$.
In the recoupling stage, we achieve precise signal reconstruction through the Adaptive Inverse Wavelet Approximation Transform (AdaIWAT), which uses similar wavelet kernels and the transpose convolution~\cite{shen2023adaptive} with adaptive weights, ensuring minimal information loss while preserving task-relevant features.

\subsection{Spatial-Frequency Mamba Block}~\label{sec:mamba}
\subsubsection{Basic Principle}
The Mamba architecture~\cite{mamba}, leveraging its State Space Model (SSM), demonstrates exceptional performance in feature modeling and MMIF tasks. Its core efficiently models the relationships between the state vector $\textbf{X}$, the state transition matrix $\textbf{A}$, the input matrix $\textbf{B}$, and the output matrix $\textbf{C}$ through linear layers, generating the final output $\textbf{Y}$. This approach exhibits high computational efficiency and global information representation capability. The basic form is a continuous-time system:
\begin{equation}
\begin{aligned}
h_{t}&=A_th_{t-1}+B_tx_t,\\
y_{t}&=C_t^\top h_t,  \label{formula4}
\end{aligned}
\end{equation}
where $x_{t}$ and $y_{t}$ are treated as scalars, representing the input and output at time \textit{t}, respectively. $h_{t}$ is the hidden state vector, and the three matrix parameters ($A_{t},B_{t},C_{t}$) can vary over \textit{t}.
On this basis, the State Space Duality (SSD) Module introduced by Mamba2~\cite{mamba2} significantly improves the global modeling capability by modeling the mapping of $\{\textbf{X}, \textbf{A}, \textbf{B}, \textbf{C}\}$ to \textbf{Y} in parallel through a single projection, similar to the Q, K, V projections in the attention mechanism.
However, despite Mamba's strong performance in multimodal tasks, its ability to adapt to high-dimensional, multimodal inputs and the frequency characteristics of 2D images remains limited. The existing SSD module is primarily designed for 1D time-series signals and struggles to handle more complex spatial-frequency features effectively.

\subsubsection{Adaptive Domain SSD}
To overcome the above limitations, we propose the Spatial-Frequency Mamba block in Fig.~\ref{fig:mamba}, which relies on a key component—Adaptive Domain SSD (AdaD-SSD)—featuring two key improvements: 1) the introduction of a spatial-aware and frequency filtering module, and 2) an update to the 2D-SSD architecture, thereby enhancing the processing capability for different frequency features.

Specifically, given the normalized fusion feature $\textbf{L}_{in}\!\in\!\mathbb{R}^{H \times W  \times  C}$, AdaD-SSD first obtains the initial feature $\textbf{L}'\!\in\!\mathbb{R}^{{HW} \times (2C'\times2G\cdot d)}$ by a linear layer projection, where $C'$ is the expansion factor, $d$ is the state dimension, and $G$ is the number of groups.
To improve the low-frequency characteristics and facilitate fusion, $\textbf{L}'$ is processed by both the spatial-aware branch $\mathcal{S}(\cdot)$ and the frequency filtering branch $\mathcal{T}(\cdot)$.

As for $\mathcal{S}(\cdot)$, the spatial convolution operator $\mathbf{W}_{se}\left ( \cdot  \right )$ with a kernel size of 3$\times$3 is combined with the \text{SiLU} activation function $\sigma(\cdot)$, to achieve the enhancement of spatial domain and dynamic feature weight adjustment, so as to strengthen the local dependence and global characterization ability of the fusion features, and obtain $\textbf{L}_{\mathcal{S}}$:
\begin{equation}
\begin{aligned}
\textbf{L}_{\mathcal{S}} = \sigma(\mathbf{W}_{se}(\mathbf{L}')) \in \mathbb{R}^{H\times W\times(C'+2G\cdot d)}.
\end{aligned}
\end{equation}

In $\mathcal{T}(\cdot)$, $\mathbf{L}'$ is first transformed to the frequency domain $\Phi$ by FFT.
To dynamically select the frequency information beneficial for fusion, we design an adaptive filter capable of learning the appropriate frequency threshold 
$\lambda$ from the parameter set $\Lambda$.
This filter efficiently optimizes the frequency components through element-wise multiplication on the 2D frequency power spectrum $\left|\Lambda\right|^2$.
Then, the updated frequency-domain features are reconstructed into $\textbf{L}_{\mathcal{T}}$ through the Inverse Fast Fourier Transform (IFFT).
This process is modeled as:
\begin{equation}
\begin{aligned}
\Lambda &= \text{FFT}(\mathbf{L}^{'}) \in \mathbb{R}^{H\times W\times(C'+2G\cdot d)},\\
\textbf{L}_{\mathcal{T}} &= \text{IFFT}(\Lambda \odot  \|\Lambda\|^2 > \lambda) \in \mathbb{R}^{H\times W\times(C'+2G\cdot d)}.
\end{aligned}
\end{equation}

Subsequently, the spatial and frequency features, $\textbf{L}_{\mathcal{S}}$ and $\textbf{L}_{\mathcal{T}}$, are fused via an adaptive cross-domain strategy.
The combined representation is then split into four components:
a state vector $\mathbf{X} \in \mathbb{R}^{{HW} \times C'}$, an input matrix $\mathbf{B} \in \mathbb{R}^{{HW} \times G \cdot d}$, an output matrix $\mathbf{C} \in \mathbb{R}^{{HW} \times G \cdot d}$, and a state transition matrix $\mathbf{A} \in \mathbb{R}^{{HW} \times C'}$:
\begin{equation}
\{ \mathbf{X},\mathbf{B},\mathbf{C},\mathbf{A}\} = \texttt{Split}(\textbf{L}_{S} \oplus 
\textbf{L}_{T}).
\end{equation}

To better model the complex spatial structure of visual inputs, AdaD-SSD extends the original 1D SSD formulation to a 2D setting.
Specifically, the image is partitioned into multiple spatial subregions, where each region maintains an independent local state. Meanwhile, global information is propagated via both intra-region and inter-region interactions.
The state update process follows a similar formulation to its 1D counterpart but incorporates spatial locality.
The updated 2D SSD equations can be rewritten from Eq.~\ref{formula4}:
\begin{equation}
\begin{aligned}
H_{ij,t} &= A_{t}H_{ij,t - 1}+B_{t}X_{ij,t},\\
Y_{ij,t} &= C_{t}^{\top}H_{ij,t},
\end{aligned}
\end{equation}
where $H_{ij,t}$ denotes the hidden state at the 2D coordinate $(i,j)$ and $t$, while $X_{ij,t}$ is the corresponding input feature at that position and time step. $A_t$ is the 2D state transition matrix, and $B_t$ is the 2D input matrix, both of which act on the hidden state and input features at the corresponding positions.

During the update process, convolution operations in 2D-SSD capture fine-grained structures within sub-regions and global interactions between regions, enabling deep fusion of spatial and frequency information. This enhances detail expressiveness while improving structural consistency, resulting in the mapping from $\{\textbf{X}, \textbf{A}, \textbf{B}, \textbf{C}\}$ to output $\textbf{Y}$:
\begin{equation}
\begin{aligned}
\textbf{Y}\!\in\!\mathbb{R}^{{HW}\times{C'}}\leftarrow \texttt{2D-SSD}\{\textbf{X}, \textbf{A}, \textbf{B}, \textbf{C}\}.
\end{aligned}
\end{equation}

Finally, the output $\textbf{L}_{out}\!\in\!\mathbb{R}^{H\times W\times C}$ of AdaD-SSD is obtained by processing the weighted result of gated activation and $\textbf{Y}$.
Notably, we retain an MLP layer~\cite{wang2024frequency} similar to the Transformer block for feature updates.

\subsection{Loss Function}
The total loss function $\mathcal{L}_{total}$ for the unified modeling of task-generalized MMIF is designed according to the configuration in~\cite{ma2022swinfusion, li2024mambadfuse}, encompassing the structural similarity (SSIM) index loss $\mathcal{L}_{ssim}$, texture loss $\mathcal{L}_{text}$, and intensity loss $\mathcal{L}_{int}$. This carefully curated combination optimizes visual fidelity while preserving intricate texture details and structural coherence across the fused multimodal inputs.

\subsubsection{SSIM Loss} 
Given that the SSIM index is a gold-standard metric for assessing image distortion in terms of luminance, contrast, and structural integrity~\cite{wang2004image}, the $\mathcal{L}_{ssim}$ term enforces structural consistency between the fused image $\mathbf{I}_{fuse}$ and the source images $\mathbf{I}_{m1}$, $\mathbf{I}_{m2}$, as:
\begin{equation}
\mathcal{L}_{ssim} =\!\sum_{i \in \{m1, m2\}} \left( 1 - \texttt{SSIM}(\mathbf{I}_{fuse}, \mathbf{I}_i) \right),
\label{formula10}
\end{equation}
where $\texttt{SSIM}(\cdot)$ computes the structural similarity between fused image and input images, ensuring that the fusion process retains both structural fidelity and perceptual quality.

\subsubsection{Texture Loss} 
A primary goal of MMIF is to effectively consolidate the intricate texture information from diverse source images into a unified fused representation.
Therefore, the texture loss $L_{text}$, as defined in Eq.~\ref{formula11}, aims to guide the network in retaining as much texture information as possible, thereby enhancing the fidelity of the fused image $\mathbf{I}_{fuse}$:
\begin{equation}
\mathcal{L}_{text} = \frac{1}{HW} |||\nabla \mathbf{I}_{fuse}| - \texttt{max}(|\nabla \mathbf{I}_{m1}|, |\nabla \mathbf{I}_{m2}|)||_1, \label{formula11}
\end{equation}
where $\nabla$ represents the Sobel gradient operator used to capture texture information, $|\cdot|$ denotes the element-wise absolute value, $||\cdot||_{1}$ is the $l_{1}$-norm, and $\texttt{max}(\cdot)$ indicates element-wise maximum selection.

\subsubsection{Intensity Loss}
A robust image fusion algorithm should generate a fused image whose intensity reflects the global intensity of the source images. To achieve this, we introduce the intensity loss $\mathcal{L}_{int}$ to guide the model in accurately capturing intensity information:
\begin{equation}
\mathcal{L}_{int} = \frac{1}{HW} ||\mathbf{I}_{fuse} -\texttt{M}(\mathbf{I}_{m1}, \mathbf{I}_{m2})||_1,
\label{formula12}
\end{equation}
where $\texttt{M}(\cdot)$ represents an element-wise aggregation operation specific to the MMIF task.

Therefore, the overall objective function $\mathcal{L}_{total}$ is defined as a weighted sum of the individual loss terms, as specified in Eq.~\ref{formula10} to Eq.~\ref{formula12}:
\begin{equation}
\mathcal{L}_{total} = \mu_1 \mathcal{L}_{ssim} + \mu_2 \mathcal{L}_{text} + \mu_3 \mathcal{L}_{int},
\end{equation}
where $\mu_{1}$=10, $\mu_{2}$=$\mu_{3}$=20 are hyperparameters that balance the contribution of each sub-loss term to the $\mathcal{L}_{total}$.


\section{Experiments and Results}

\subsection{Experimental Setup}
\subsubsection{Datasets} 
We evaluated the proposed AdaSFFuse in the current main image fusion tasks: Infrared-Visible Fusion (\textbf{IVF}), multi-exposure fusion (\textbf{MEF}), multi-focus fusion (\textbf{MFF}), and medical image fusion (\textbf{MIF}).

\noindent{\textbullet~\textbf{IVF.}} This task uses the LLVIP~\cite{zhang2023visible} dataset~\footnote{\url{https://github.com/bupt-ai-cz/LLVIP}}, which contains 12,025 training image pairs and 3,463 test image pairs.

\noindent{\textbullet~\textbf{MEF.}} The training dataset is SICE~\cite{Cai2018deep}\footnote{\url{https://github.com/csjcai/SICE}} that consists of 542 pairs of images. The evaluation dataset is MEFB~\cite{zhang2021benchmarking}\footnote{\url{https://github.com/xingchenzhang/MEFB}}, which contains 100 pairs of images with various scenes.

\noindent{\textbullet~\textbf{MFF.}} The training images includes 710 pairs of images from Real-MFF~\cite{zhang2020real}\footnote{\url{https://github.com/Zancelot/Real-MFF}} and 120 pairs of images from MFI-WHU~\cite{zhang2021mff}\footnote{\url{https://github.com/HaoZhang1018/MFI-WHU}}. The evaluation is conducted on 20 pairs of images from Lytro\cite{nejati2015multi}\footnote{\url{https://mansournejati.ece.iut.ac.ir/content/lytro-multi-focus-dataset}} and 13 pairs of images from MFFW~\cite{xu2020mffw}.

\noindent{\textbullet~\textbf{MIF.}}
This task is trained and evaluated on the public images from the Harvard Medical website\footnote{\url{https://www.med.harvard.edu/AANLIB/home.html}}. Namely, 160 training pairs and  24 test pairs in the CT-MRI dataset; 245 training pairs and 24 test pairs in the PET-MRI dataset; and 333 training pairs and 24 test pairs in the SPECT-MRI dataset, totaling 738 training image pairs and 72 test image pairs. 


\subsubsection{Evaluation Metrics} 
Following the common practice in previous work~\cite{sun2024deep,chen2024frequency,cheng2023mufusion}, 
we use eight metrics to quantitatively measure the fusion results: entropy (EN)\cite{roberts2008assessment}, standard deviation (SD)\cite{rao1997fibre}, spatial frequency (SF)\cite{eskicioglu1995image}, mutual information (MI)\cite{haghighat2011non}, the sum of difference correlation (SCD)\cite{qu2002information}, visual information fidelity (VIF)\cite{han2013new}, Qabf~\cite{xydeas2000objective}, and structural similarity (SSIM)~\cite{ma2015perceptual} index metric. 

\subsection{Implementation Details}
In the preprocessing stage, all images were converted into single-channel inputs following the training settings of~\cite{li2024mambadfuse,ma2022swinfusion}. The images were then organized into 128$\times$128 pixel patches and normalized to serve as training data for all methods. 
During the training phase, the $N_{1}$ and $N_{2}$ parameters of AdaSFFuse were set to 2 and 4, respectively, with a model depth $C$ = 64 and an MLP structure with an expansion ratio of 2. We also use the Adam optimizer with an initial learning rate of \(1 \times 10^{-4}\), and the batch size of 120.
In the testing phase, RGB images are first converted to the YCbCr color space, with the Y (luminance) channel used as input for the fusion model, as it contains the primary structural and intensity information. Infrared images, PET images, \etc, are used directly as model inputs. Finally, the Cb and Cr (blue and red chrominance) channel information is mapped back to the RGB color space along with the fused image.

\begin{table*}[t!]
\tabcolsep 2pt
\renewcommand\arraystretch{1.1}
\caption{Quantitative comparison results of the Infrared-Visible Image Fusion (IVF) and Multi-Exposure Image Fusion (MEF) tasks, where the best and second-best values are indicated in {\color{red}\textbf{red}} and {\color{blue}\textbf{blue}}, respectively.
}
\resizebox{1.0\linewidth}{!}{
\begin{tabular}{>{\raggedleft\arraybackslash}p{2.0cm}>{\raggedright\arraybackslash}p{0.8cm}||cccccccc|cccccccc}
\hline
\thickhline
\rowcolor[HTML]{f8f9fa}
 & & \multicolumn{8}{c|}{Infrared-Visible Image Fusion (IVF)} & \multicolumn{8}{c}{Multi-Exposure Image Fusion (MEF)}  \\
\rowcolor[HTML]{f8f9fa}
\multirow{-2}{*}{Method} & & EN $\uparrow$ & SD $\uparrow$  & SF $\uparrow$ & MI $\uparrow$ & SCD $\uparrow$  & VIF $\uparrow$  & Qabf $\uparrow$ & SSIM $\uparrow$ & EN $\uparrow$ & SD $\uparrow$  & SF $\uparrow$  & MI $\uparrow$ & SCD $\uparrow$  & VIF $\uparrow$  & Qabf $\uparrow$ & SSIM $\uparrow$ \\ \hline
DRDEC~\cite{sun2024deep}\!\!&\!\!\pub{TMM2024} & 6.41 & 35.99 & 10.36 & 2.10 & 0.83 & 0.69 & 0.46 & 1.42 & 6.87 & 52.58 & 16.57 & 4.38 & 0.47 & 1.33 & 0.66 & 1.23 \\
HitFusion~\cite{chen2024hitfusion}\!\!&\!\!\pub{TMM2024}  & 6.43 & 36.84 & 10.12 & 1.98 & 0.92 & 0.59 & 0.40 & 1.41 & 6.87 & 51.19 & 16.63 & 4.06 & 0.55 & 1.27 & 0.63 & 1.22 \\
EMMA~\cite{Zhao_2024_CVPR}\!\!&\!\!\pub{CVPR2024} & 7.25& 41.54& 11.39& 2.53& 1.23& 0.60& 0.67& {\color{blue}1.44} & 6.75 & 60.55 & {\color{blue}20.80} & 4.68 & 0.72 & 1.09 & 0.54 & 1.12 \\
ITFuse~\cite{tang2024itfuse}\!\!&\!\!\pub{PR2024} & 6.78 & 33.41 & 11.71 & 1.44 & 1.08 & 0.27 & 0.24 & 1.00 & 6.95 & 59.47 & 14.87 & 4.21 & 0.53 & 1.04 & 0.41 & 1.03 \\
MUFusion~\cite{cheng2023mufusion}\!\!&\!\!\pub{IF2023} & 5.59 & 28.21 & 11.50 & 1.33 & 0.83 & 0.51 & 0.36 & 1.30 & 6.44 & 52.48 & 16.46 & 4.34 & 0.54 & 1.05 & 0.54 & 1.13 \\
DDFM~\cite{zhao2023ddfm}\!\!&\!\!\pub{ICCV2023} & 7.06 & 39.80 & 13.17 & 2.87 & 1.45 & 0.61 & 0.46 & 1.41 & 7.03 & 55.05 & 16.33 & 4.77 & 0.65 & 1.28 & 0.55 & 1.16 \\
MambaDFuse~\cite{li2024mambadfuse}\!\!&\!\!\pub{arXiv2024} & 7.32& 47.25& 13.73& 2.60& {\color{blue}1.46}& 0.91& 0.56& 1.43 & 6.30& 61.40& 17.67& {\color{blue}4.94}& 0.63& 1.42& 0.69& 1.23 \\
CDDFuse~\cite{zhao2023cddfuse}\!\!&\!\!\pub{CVPR2023}  & {\color{blue} 7.34}& 47.07& 14.91& 2.77& 1.46& {\color{blue} 1.01}& {\color{blue} 0.74}& 1.44 & 7.04& {\color{blue}64.54}& 20.05& 4.91& {\color{blue}0.73}& {\color{blue}1.46}& {\color{blue}0.75}& {\color{blue}1.29} \\
SwinFusion~\cite{ma2022swinfusion}\!\!&\!\!\pub{JAS2022} & 7.33&  {\color{blue}47.27}&  {\color{blue}14.92}&  {\color{blue}2.88}& 1.46& 0.94& 0.69& 1.42 & {\color{blue}7.07}& 60.53& 18.81& 4.72& 0.62& 1.43& {\color{red}0.76}& 1.21  \\ 
SHIP~\cite{zheng2024probing}\!\!&\!\!\pub{CVPR2024} & 6.74& 32.01 &12.73 & 1.91 & 0.93 & 0.74& 0.61& 1.25 &6.73 & 50.11 & 17.15& 4.68& 0.68& 1.15& 0.65& 1.15 \\ \hline

\rowcolor{gray!18}
\multicolumn{2}{c||}{\textbf{AdaSFFuse (Ours)}} & {\color{red}\textbf{7.39}}& {\color{red}\textbf{47.60}}& {\color{red}\textbf{15.28}}& {\color{red}\textbf{3.06}}& {\color{red}\textbf{1.48}}& {\color{red}\textbf{1.01}}& {\color{red}\textbf{0.79}}& {\color{red}\textbf{1.51}} & {\color{red}\textbf{7.28}}& {\color{red}\textbf{66.40}}& {\color{red}\textbf{21.93}}& {\color{red}\textbf{5.37}}& {\color{red}\textbf{0.76}}& {\color{red}\textbf{1.47}}& {\color{red}\textbf{0.76}}& {\color{red}\textbf{1.32}}  \\ \hline
\end{tabular}}
\label{tab:results_IVF_mef}
\end{table*}
\begin{figure*}[t!]
\centering
\begin{overpic}[width=1.0\linewidth]{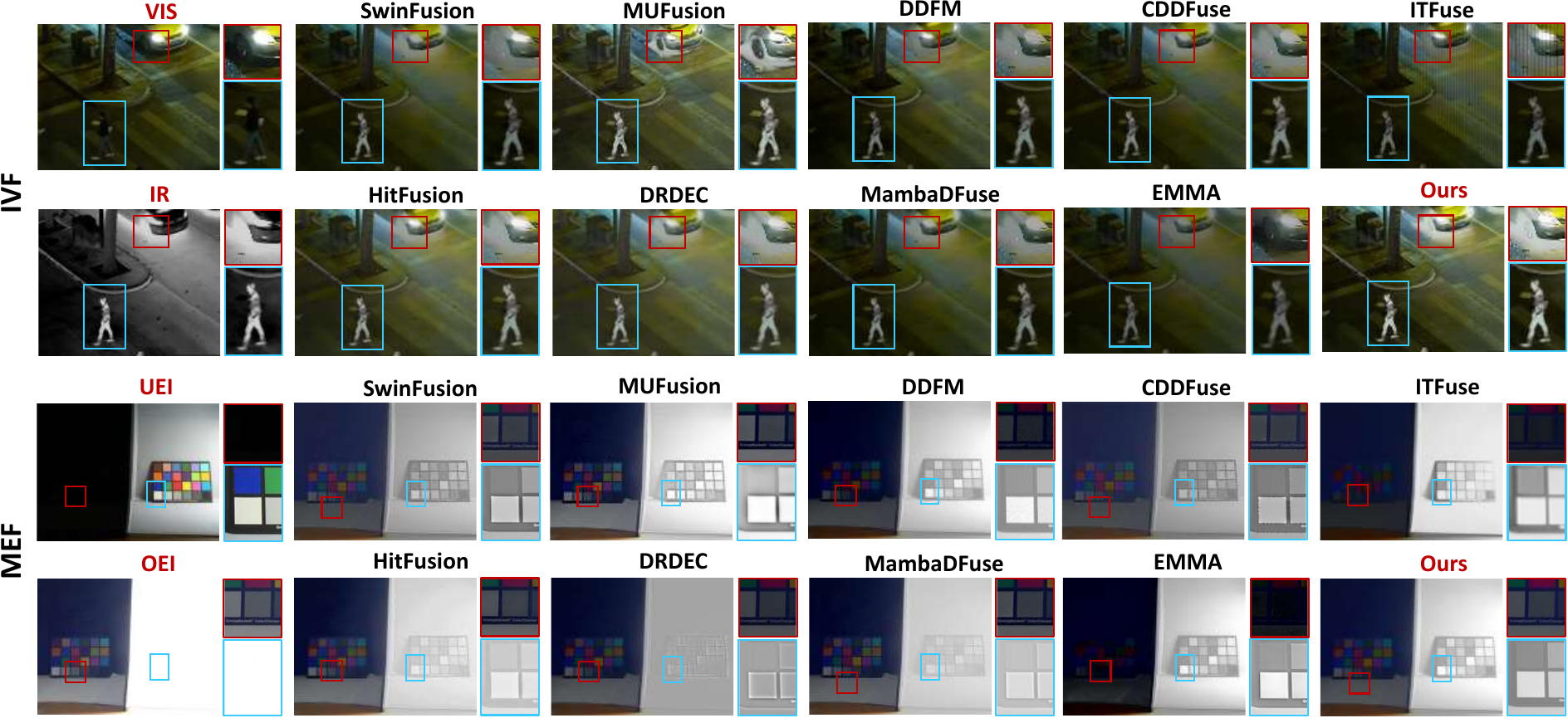}
\put(30,21){\small{\cite{ma2022swinfusion}}}
\put(30,45){\small{\cite{ma2022swinfusion}}}

\put(46,21){\small{\cite{cheng2023mufusion}}}
\put(46,45){\small{\cite{cheng2023mufusion}}}

\put(61.5,21){\small{\cite{zhao2023ddfm}}}
\put(61.5,45){\small{\cite{zhao2023ddfm}}}

\put(78.5,21){\small{\cite{zhao2023cddfuse}}}
\put(78.5,45){\small{\cite{zhao2023cddfuse}}}

\put(94,21){\small{\cite{tang2024itfuse}}}
\put(94,45){\small{\cite{tang2024itfuse}}}

\put(30,9.6){\small{\cite{chen2024hitfusion}}}
\put(30,33){\small{\cite{chen2024hitfusion}}}

\put(46,9.6){\small{\cite{sun2024deep}}}
\put(46,33){\small{\cite{sun2024deep}}}

\put(63.5,9.6){\small{\cite{li2024mambadfuse}}}
\put(63.5,33){\small{\cite{li2024mambadfuse}}}

\put(78,9.6){\small{\cite{Zhao_2024_CVPR}}}
\put(78,33){\small{\cite{Zhao_2024_CVPR}}}
\end{overpic}
\caption{Qualitative comparison of the Infrared-Visible Image Fusion (IVF) and Multi-Exposure Image Fusion (MEF) tasks. To provide a clearer comparison, we highlighted two competitive regions ({\color{red}\textbf{red}} and {\color{blue}\textbf{blue}} boxes) in each image and zoomed in on the right to visualize the performance difference.}
\label{fig:vis_IVF_MEF}
\end{figure*}
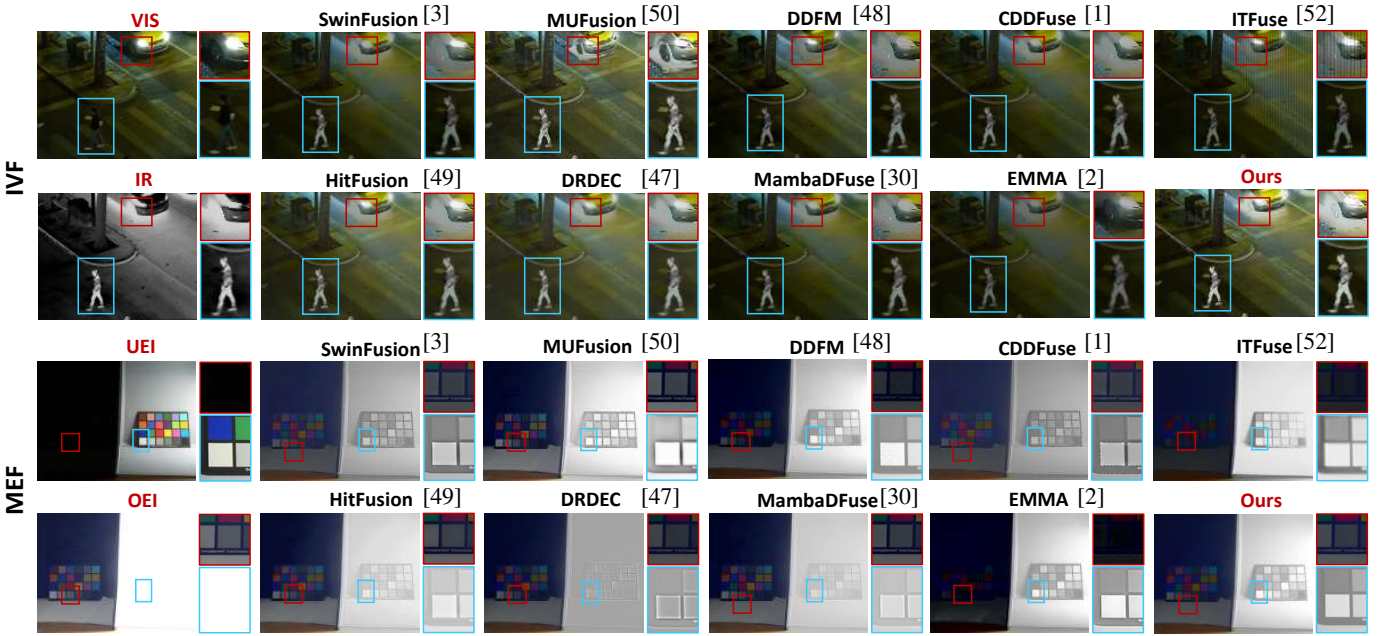

\subsection{Comparison with State-of-the-Arts}
We compare the proposed AdaSFFuse with ten SOTA MMIF methods.
These include one residual network-based method: 
DRDEC~\cite{sun2024deep}, which uses a deep residual network combined with a densely connected encoder-decoder architecture; one generative model-based method: HitFusion~\cite{chen2024hitfusion}, which employs a matching-type algorithm to generate fusion results through feature-point matching; two methods based on attentional mechanisms: MUFusion~\cite{cheng2023mufusion} and ITFuse~\cite{tang2024itfuse}, which utilize a variety of transform domains and attentional mechanisms for image fusion to improve the multimodal image fusion effect; three multi-scale advanced modular architectures (Transformer, Mamba and Diffusion Model): SwinFusion~\cite{ma2022swinfusion}, MambaDFuse~\cite{li2024mambadfuse}, DDFM~\cite{zhao2023ddfm} and EMMA~\cite{Zhao_2024_CVPR}, combining multi-scale feature extraction and dense connectivity for image fusion; a high-order interactive collaborative mechanism: SHIP~\cite{zheng2024probing} integrates spatial and channel dimensions; and Transformer-based two-stage training method: CDDFuse~\cite{zhao2023cddfuse}.

\begin{table*}[t!]
\tabcolsep 2pt
\renewcommand\arraystretch{1.1}
\caption{Quantitative comparison results of the Multi-Focus Image Fusion (MFF) and Medical Image Fusion (MIF) tasks, where the best and second-best values are indicated in {\color{red}\textbf{red}} and {\color{blue}\textbf{blue}}, respectively.
}
\resizebox{1.0\linewidth}{!}{
\begin{tabular}{>{\raggedleft\arraybackslash}p{2.0cm}>{\raggedright\arraybackslash}p{0.8cm}||cccccccc|cccccccc}
\hline
\thickhline
\rowcolor[HTML]{f8f9fa}
 & & \multicolumn{8}{c|}{Multi-Focus Image Fusion (MFF)} & \multicolumn{8}{c}{Medical Image Fusion (MIF)}  \\
\rowcolor[HTML]{f8f9fa}
\multirow{-2}{*}{Method} & & EN $\uparrow$ & SD $\uparrow$  & SF $\uparrow$ & MI $\uparrow$ & SCD $\uparrow$  & VIF $\uparrow$  & Qabf $\uparrow$ & SSIM $\uparrow$ & EN $\uparrow$ & SD $\uparrow$  & SF $\uparrow$  & MI $\uparrow$ & SCD $\uparrow$  & VIF $\uparrow$  & Qabf $\uparrow$ & SSIM $\uparrow$ \\ \hline
DRDEC~\cite{sun2024deep}\!\!&\!\!\pub{TMM2024} & 7.39& 56.11& 11.59& 4.33& 0.66& 1.03& 0.54& 1.65 & 4.17& 58.92& 15.72& 1.97& 1.06& 0.52& 0.58& 1.06\\
HitFusion~\cite{chen2024hitfusion}\!\!&\!\!\pub{TMM2024}  & 7.39& 55.97& 12.55& 4.34& 0.71& 1.01& 0.48& 1.68 & 4.20& 60.30& 19.84& 1.89& 1.19& 0.49& 0.45& 1.07\\
EMMA~\cite{Zhao_2024_CVPR}\!\!&\!\!\pub{CVPR2024} & 6.53& 51.54& 13.39& 4.03& 0.53& 0.89& 0.47& 1.50 & 5.00& 67.57& 25.88& 2.10& 1.06& 0.58& 0.58& 1.39\\
ITFuse~\cite{tang2024itfuse}\!\!&\!\!\pub{PR2024} & 7.08& 51.36& 12.09& 4.39& 0.51& 0.86& 0.53& 1.27 & 4.95& 72.66& 23.79& 1.94& 1.39& 0.49& 0.44& 1.29 \\
MUFusion~\cite{cheng2023mufusion}\!\!&\!\!\pub{IF2023} & 6.79& 55.64& 18.6& 4.37& 0.62& 1.22& 0.55& 1.22 & 5.07& 79.13& 30.14& 2.48& 1.55& 0.65& 0.64& 1.28 \\
DDFM~\cite{zhao2023ddfm}\!\!&\!\!\pub{ICCV2023} & 6.95& 53.36& 18.74& 4.11& 0.45& 1.05& 0.59& 1.22 & 4.88& 65.34& 29.36& 2.48& 1.31& 0.71& 0.69& 1.41 \\
MambaDFuse~\cite{li2024mambadfuse}\!\!&\!\!\pub{arXiv2024} & 7.39& 56.39& 19.26& 4.43& 0.69& 1.13& 0.71& {\color{blue}1.71} & 4.98& {\color{blue}79.87}& {\color{blue}30.65}& 2.41& {\color{blue}1.58}& 0.71& 0.70& 1.45  \\
CDDFuse~\cite{zhao2023cddfuse}\!\!&\!\!\pub{CVPR2023}  & {\color{blue}7.42}& {\color{red}56.59}& {\color{blue}20.36}& 4.70& 0.61& 1.23& 0.72& 1.38 & {\color{red}5.08}& 75.89& 30.12& {\color{red}2.57}& 1.48& {\color{blue}0.75}& {\color{blue}0.72}& 1.32  \\
SwinFusion~\cite{ma2022swinfusion}\!\!&\!\!\pub{JAS2022} & 7.39& 56.30& 20.03& {\color{blue}4.73}& 0.58& {\color{blue}1.23}& {\color{red}0.74}& 1.69 & 4.96& 79.45& 30.42& 2.46& 1.54& 0.74& 0.71& {\color{blue}1.45} \\ 
SHIP~\cite{zheng2024probing}\!\!&\!\!\pub{CVPR2024} & 7.37& 55.00 & 12.94 & 4.30& 0.56& 1.11& 0.69&1.67& 5.03& 58.00&25.63& 2.15& 1.14& 0.63& 0.64& 1.31 \\ \hline
\hline
\rowcolor{gray!18}
\multicolumn{2}{c||}{\textbf{AdaSFFuse (Ours)}} & {\color{red}\textbf{7.48}}& {\color{blue}\textbf{56.53}}& {\color{red}\textbf{20.44}}& {\color{red}\textbf{4.87}}& {\color{red}\textbf{0.75}}& {\color{red}\textbf{1.26}}& {\color{blue}\textbf{0.72}}& {\color{red}\textbf{1.78}} & {\color{blue}\textbf{5.07}}& {\color{red}\textbf{79.96}}& {\color{red}\textbf{30.81}}& {\color{blue}\textbf{2.49}}& {\color{red}\textbf{1.59}}& {\color{red}\textbf{0.80}}& {\color{red}\textbf{0.74}}& {\color{red}\textbf{1.49}} \\ \hline
\end{tabular}}
\label{tab:results_mef_mif}
\end{table*}
\begin{figure*}[t!]
\centering
\begin{overpic}[width=1.0\linewidth]{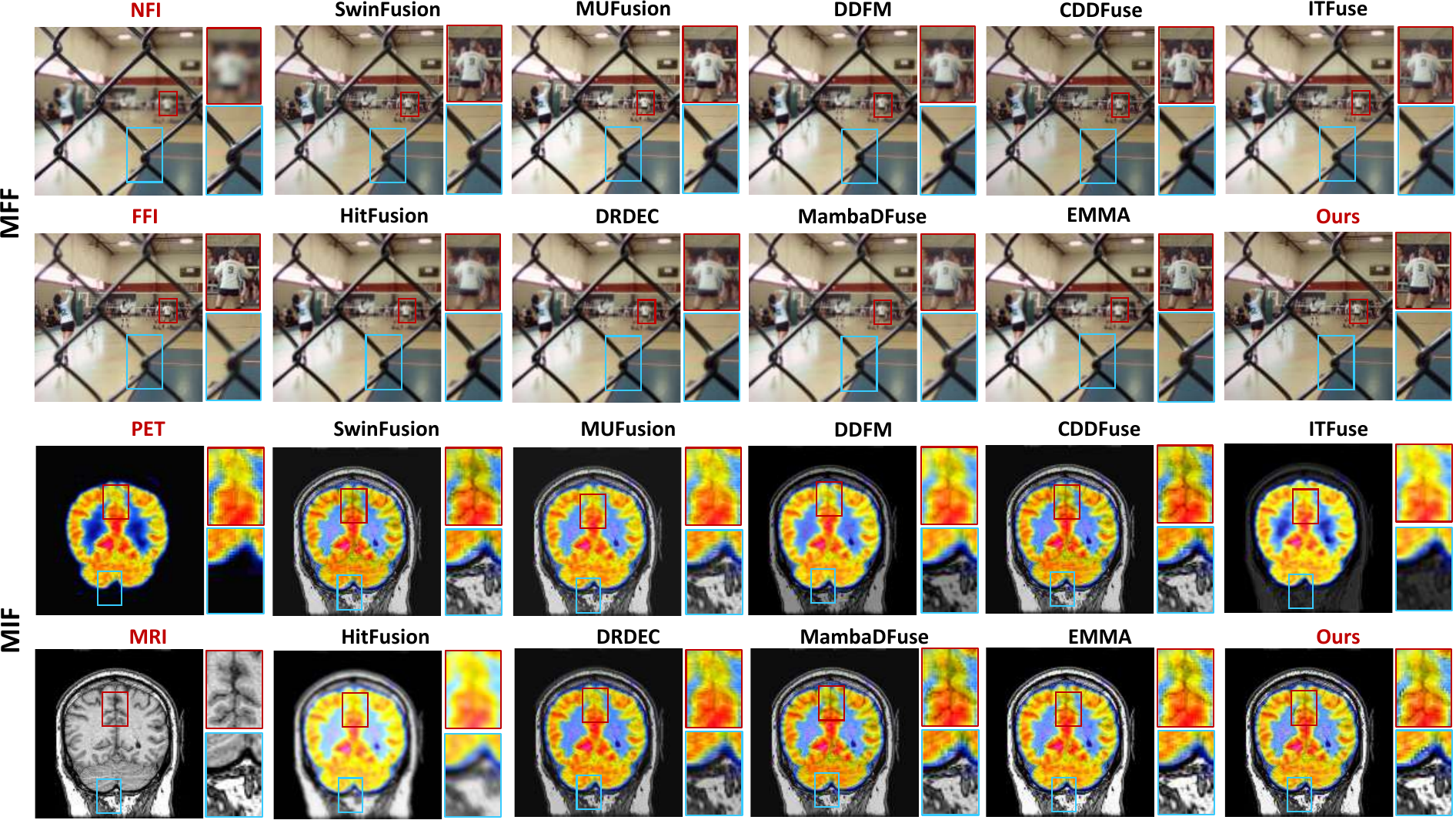}
\put(30,26.5){\small{\cite{ma2022swinfusion}}}
\put(30,55.3){\small{\cite{ma2022swinfusion}}}

\put(46,26.5){\small{\cite{cheng2023mufusion}}}
\put(46,55.3){\small{\cite{cheng2023mufusion}}}

\put(64,26.5){\small{\cite{zhao2023ddfm}}}
\put(64,55.3){\small{\cite{zhao2023ddfm}}}

\put(78,26.5){\small{\cite{zhao2023cddfuse}}}
\put(78,55.3){\small{\cite{zhao2023cddfuse}}}

\put(94,26.5){\small{\cite{tang2024itfuse}}}
\put(94,55.3){\small{\cite{tang2024itfuse}}}

\put(30,12.2){\small{\cite{chen2024hitfusion}}}
\put(30,41){\small{\cite{chen2024hitfusion}}}

\put(46,12.2){\small{\cite{sun2024deep}}}
\put(46,41){\small{\cite{sun2024deep}}}

\put(64,12.2){\small{\cite{li2024mambadfuse}}}
\put(64,41){\small{\cite{li2024mambadfuse}}}

\put(78,12.2){\small{\cite{Zhao_2024_CVPR}}}
\put(78,41){\small{\cite{Zhao_2024_CVPR}}}

\end{overpic}
\caption{Qualitative comparison of the Multi-Focus Image Fusion (MFF) and Medical Image Fusion (MIF) tasks. To provide a clearer comparison, we highlighted two competitive regions ({\color{red}\textbf{red}} and {\color{blue}\textbf{blue}} boxes) in each image and zoomed in on the right to visualize the performance difference.}
\label{fig:vis_MFF_MIF}
\end{figure*}
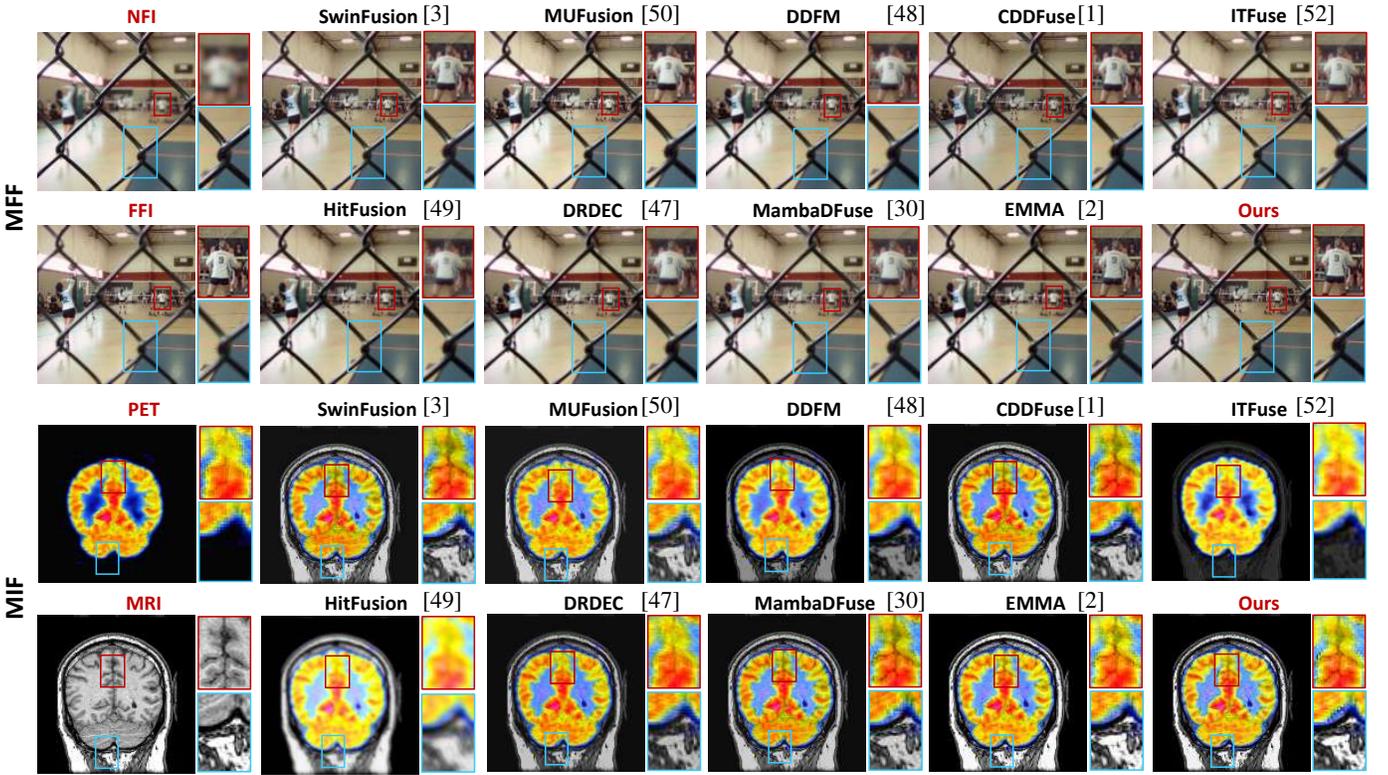

\subsubsection{Visible and Infrared Image Fusion}
As shown in Table~\ref{tab:results_IVF_mef}, the proposed AdaSFFuse outperforms previous methods on all metrics, especially on key metrics such as MI, Qabf, and SSIM. These results indicate that the information transfer from the source image to the fused image is extremely efficient while maintaining a low level of image distortion, which fully verifies the effectiveness and superiority of the method.
These excellent quantitative results are attributed to the organic combination of feature extraction, learnable wavelet transform, multilayer perception, and frequency fusion modules in the method.
Additionally, we also give the qualitative comparison in Fig.~\ref{fig:vis_IVF_MEF}. 
In low-light environments, AdaSFFuse significantly improves the recognizability of target objects and clearly distinguishes the foreground from the background.
In the application of road scenes, AdaSFFuse enhances the rendering effect of VIS images by fusing IR images on the Y-channel, while completely retaining the chromaticity information of VIS images, thus realizing a superior scene representation.

\subsubsection{Multi-Exposure Image Fusion}
In this section, we evaluate the performance of the proposed AdaSFFuse in the task of Multi-Exposure Image Fusion (MEF). The quantitative results are presented in Table~\ref{tab:results_IVF_mef}. These results show that AdaSFFuse achieves the best average performance across all evaluation metrics, compared to the SOTA method CDDFuse~\cite{zhao2023cddfuse}, particularly in SD, SF, and SSIM, where it reaches scores of 66.4, 21.93, and 1.32, respectively. This indicates that the fused images not only preserve more of the original image characteristics but also exhibit improved structural coherence and intensity preservation. Figure~\ref{fig:vis_IVF_MEF} illustrates the fusion results on the MEF dataset using various methods. Compared to the other SOTA methods, our approach demonstrates superior visual fidelity in color palette preservation, with enhanced detail clarity, contrast, and brightness balance, while also offering better perceptual enhancement of local details.
\begin{figure*}[t!]
\centering
\includegraphics[width=1.0\linewidth]{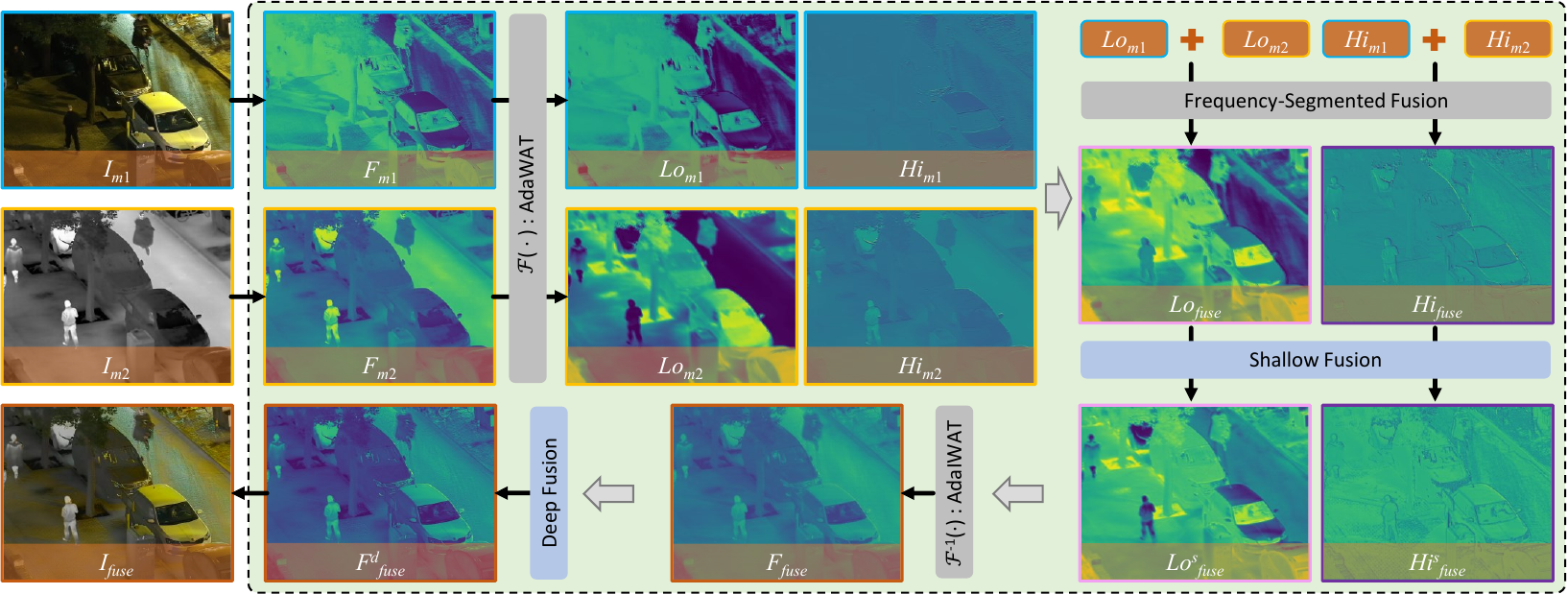}
\caption{Visualization of feature maps in AdaSFFuse on the LLVIP dataset for the IVF task. $\textbf{Lo}_{m1}$ and $\textbf{Lo}_{m2}$ capture low-frequency structures from the VIS and IR images, respectively, while $\textbf{Hi}_{m1}$ and $\textbf{Hi}_{m2}$ highlight high-frequency details. After Frequency-Segmented Fusion and shallow fusion by Spatial-Frequency Mamba, the fused low- and high-frequency maps $\textbf{Lo}^{s}_{fuse}$ and $\textbf{Hi}^{s}_{fuse}$ combine more complete information. The final fused image, $\textbf{I}_{fuse}$, preserves clear textures and shapes from both modalities.} 
\label{Fig:vir}
\end{figure*}
\begin{figure}[t!]
\centering
\includegraphics[width=1\linewidth]{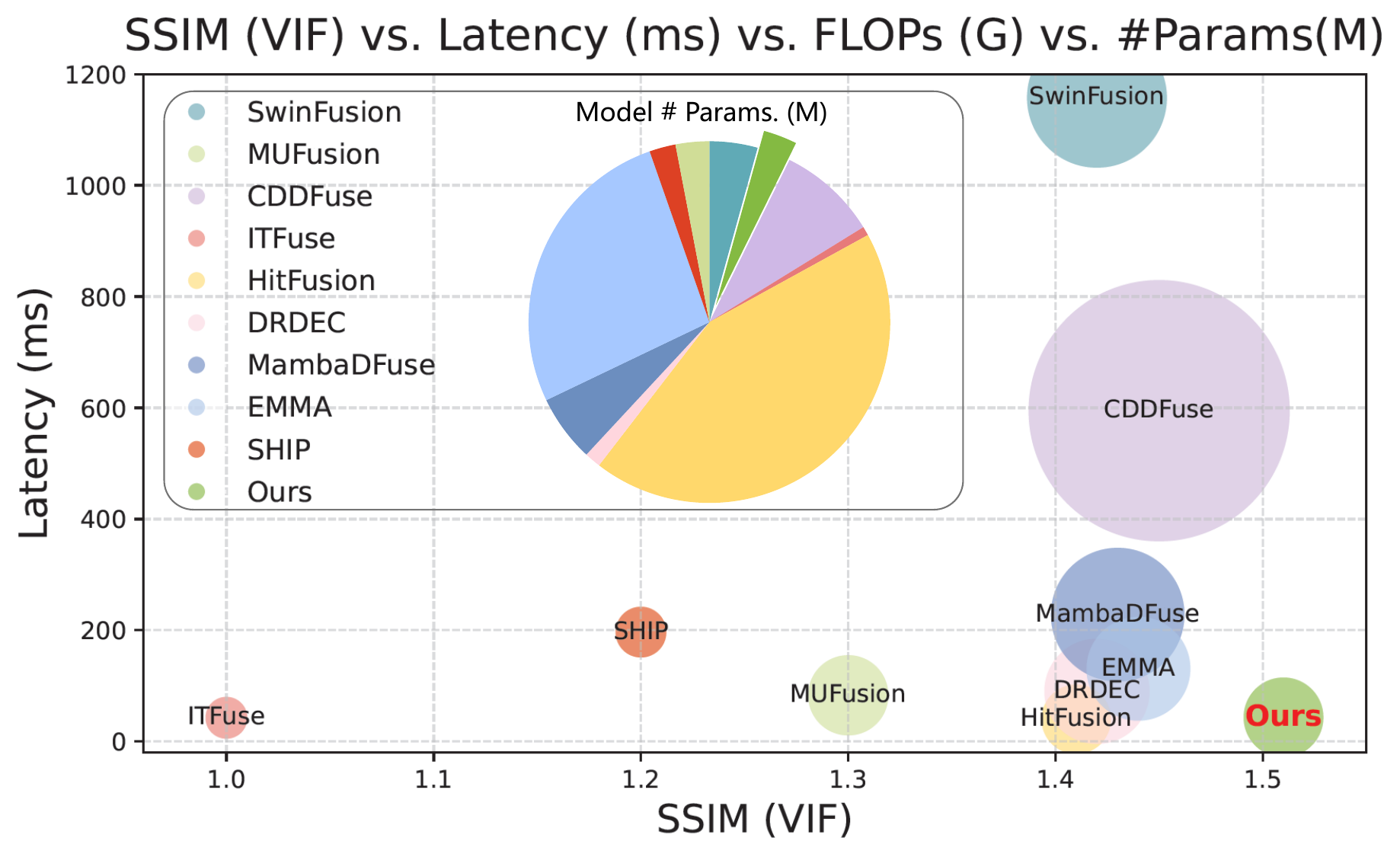}
\caption{Comparison of model complexity analysis with SOTA methods. We compared the model size, complexity, and latency of AdaSFFuse with other SOTA methods at a resolution of 512$\times$512. The results demonstrate that we achieve superior performance within an acceptable range of parameters.}
\label{fig:results_flops}
\end{figure}

\subsubsection{Multi-focus Image Fusion}
As shown in Table~\ref{tab:results_mef_mif}, we also evaluate AdaSFFuse on Multi-focus Image Fusion (MIF).
The higher MI, SSIM, and VIF scores validate AdaSFFuse's exceptional ability to preserve information fidelity and structural consistency.
The superior performance in SF and SCD highlights its strength in maintaining spatial details and enhancing spatial correlations, ensuring edge and texture integrity.
Moreover, the high SD and Qabf scores further affirm AdaSFFuse's capability to enhance contrast and overall fusion quality. The highest EN score reflects its ability to effectively integrate rich information from source images. These quantitative results demonstrate the robustness and versatility of AdaSFFuse across various fusion tasks. Qualitative analysis in Fig.~\ref{fig:vis_MFF_MIF}, comparing AdaSFFuse with nine SOTA methods on the Lytro and MFFW datasets, shows its clear advantage. While other methods exhibit issues such as blurred edges (DDFM, MUFusion), excessive focus on distant objects (DRDEC), or feature loss (EMMA, ITFuse), AdaSFFuse preserves image details, reduces boundary blurring, and enhances the visual quality of the fused image. Despite minor detail loss in the far-focus image, AdaSFFuse outperforms other methods in terms of brightness, color fidelity, and contrast, achieving the best fusion results. These results demonstrate AdaSFFuse’s superior performance and reliability in image fusion tasks.

\subsubsection{Medical Image Fusion}
Table~\ref{tab:results_mef_mif} presents the quantitative results on Medical Image Fusion (MIF), where AdaSFFuse demonstrates highly competitive performance. In particular, it outperforms the best model CDDFuse~\cite{zhao2023cddfuse} in VIF, Qabf, and SSIM metrics across the CT-MRI, PET-MRI, and SPECT-MRI datasets, achieving scores of 0.80, 0.74, and 1.49, respectively.
Figure~\ref{fig:vis_MFF_MIF} shows a comparison of fusion results between AdaSFFuse and existing methods on a pair of PET-MRI images. The fused image generated by AdaSFFuse exhibits enhanced contrast and luminance, offering improved differentiation between various tissues and lesion regions.
Furthermore, the fused image preserves more fine structures and details from the original PET and MRI images, demonstrating the exceptional capability of our method in retaining intricate details and improving overall image quality.

\subsection{Feature Interpretive Analysis}
To validate the contribution of critical modules in the designed AdaSFFuse, we conduct an in-depth analysis of the feature maps across the different modules.
As discussed in Sec.~\ref{adawat}, AdaWAT is capable of adaptively capturing both low- and high-frequency features from multimodal images. In Figure~\ref{Fig:vir}, $\textbf{Lo}_{m1}$ captures the low-frequency visible light shapes from the VIS image, while $\textbf{Lo}_{m2}$ focuses on the low-frequency structures of dominant targets (such as pedestrians and vehicles) in the IR image. The high-frequency components, $\textbf{Hi}_{m1}$ and $\textbf{Hi}_{m1}$, emphasize guiding features like textures and contours (\eg, tile patterns and target outlines) from each modality. After Frequency-Segmented Fusion and the shallow fusion by the Spatial-Frequency Mamba, the low- and high-frequency fused maps, $\textbf{Hi}^{s}_{fuse}$ and $\textbf{Lo}^{s}_{fuse}$, contain more complete fusion information.
Following AdaIWAT and deep fusion, the final fusion map, $\textbf{F}_{fuse}$, captures clear textures and shapes from both modalities, resulting in a high-quality fused image, $\textbf{I}_{fuse}$.
These results demonstrate that the proposed AdaSFFuse effectively integrates cross-domain information from multiple modalities, progressively aligning them to enhance model performance.

\subsection{Computational Complexity Analysis}
To comprehensively evaluate the computational efficiency of the proposed method, we conduct a systematic quantitative analysis of computational complexity.
The results are shown in Fig.~\ref{fig:results_flops}. 
(1) In terms of the number of model parameters, the proposed method only requires 0.78 M parameters to achieve an excellent fusion effect, which is a significant compression of the number of parameters compared with the existing methods such as HitFusion (9.81 M) and EMMA (1.52 M), and the reduction rate reaches 92.0$\%$ and 48.7$\%$, respectively.
Although slightly higher compared to ITFuse (0.08 M), this slight increase in the number of parameters is an acceptable trade-off considering the excellent performance of the present method in all performance metrics.
(2) For computational complexity, the FLOPs of the present method are 82.81 G, which are 67.5$\%$ and 64.4$\%$ lower than those of SwinFusion (254.92 G) and MambaDFuse (232.60 G), respectively. These results demonstrate the significant advantage of the present method in computational efficiency.
(3) In terms of inference speed, the average latency of the proposed method is 70.13 ms, which is higher than that of some lightweight methods, but still demonstrates a significant speed advantage over computationally intensive methods such as SwinFusion (1157.83 ms) and MambaDFuse (227.97 ms), realizing a time saving of 93.9$\%$ and 69.2$\%$, respectively.
This result fully demonstrates the efficiency of the proposed method in practical application scenarios.
The comprehensive analysis shows that the proposed method achieves excellent computational efficiency while maintaining a low number of parameters through careful network design and reaches an ideal balance between model complexity and performance.
\begin{table}[t!]
\renewcommand\arraystretch{1.2}
\tabcolsep 2pt
\caption{Ablation Results for Key Components. Here, we use the network that removes AdaWAT, the Spatial-Frequency Mamba with shallow fusion $\textbf{M}_{shallow}$, and the Spatial-Frequency Mamba with deep fusion $\textbf{M}_{deep}$ as the baseline.}
\resizebox{1.0\linewidth}{!}{
\begin{tabular}{ll@{\hspace{4pt}}||cccccccc}
\hline
\thickhline
\rowcolor[HTML]{f8f9fa} & Networks & EN$\uparrow$ & SD$\uparrow$ & SF$\uparrow$ & MI$\uparrow$ & SCD$\uparrow$ & VIF$\uparrow$ & Qabf$\uparrow$ & SSIM$\uparrow$ \\
\hline
A.I & Baseline & 4.58 & 27.69 & 9.86 & 0.83 & 0.65 & 0.21 & 0.27 & 0.89 \\
A.II & $w$/ AdaWAT & 6.47 & 32.59 & 11.01 & 1.23 & 0.79 & 0.44 & 0.49 & 1.04 \\
A.III & $w$/ $\textbf{M}_{shallow}$ & {\color{blue}7.25} & {\color{blue}43.58} & {\color{blue}14.69} & {\color{blue}2.63} & {\color{blue}1.40} & {\color{blue}0.88} & {\color{blue}0.69}& {\color{blue}1.41} \\
\rowcolor{gray!10}B.IV & $w$/ $\textbf{M}_{deep}$\!\!& {\color{red}\textbf{7.39}}& {\color{red}\textbf{47.60}}& {\color{red}\textbf{15.28}}& {\color{red}\textbf{3.06}}& {\color{red}\textbf{1.48}}& {\color{red}\textbf{1.01}}& {\color{red}\textbf{0.79}}& {\color{red}\textbf{1.51}}\\
\hline
\end{tabular}
}
\label{tab:module}
\end{table}

\begin{table}[t!]
\renewcommand\arraystretch{1.2}
\tabcolsep 2pt
\caption{Ablation Results for Frequency Decoupling Methods.}
\resizebox{1.0\linewidth}{!}{
\begin{tabular}{ll@{\hspace{4pt}}||cccccccc}
\hline\thickhline
\rowcolor[HTML]{f8f9fa} & Configs & EN$\uparrow$ & SD$\uparrow$ & SF$\uparrow$ & MI$\uparrow$ & SCD$\uparrow$ & VIF$\uparrow$ & Qabf$\uparrow$ & SSIM$\uparrow$ \\
\hline
B.I & FFT~\cite{shen2024spatial} & 6.89 & 39.28 & 11.87 & 2.51 & 1.26 & 0.69 & 0.65 & 1.38\\
B.II & LapF~\cite{iqbal2020underwater}  & 6.94 & 41.23 & 13.23 & 2.49 & 1.35 & 0.75 & 0.58 & 1.33 \\
B.III & WAT~\cite{huang2022winnet}   & {\color{blue}7.21} & {\color{blue}44.88} & {\color{blue}14.65} & {\color{blue}2.61} & {\color{blue}1.37} & {\color{blue}0.81} & {\color{blue}0.66} &{\color{blue} 1.40} \\
\rowcolor{gray!10}B.IV & \textbf{AdaWAT(Ours)}\!\!& {\color{red}\textbf{7.39}}& {\color{red}\textbf{47.60}}& {\color{red}\textbf{15.28}}& {\color{red}\textbf{3.06}}& {\color{red}\textbf{1.48}}& {\color{red}\textbf{1.01}}& {\color{red}\textbf{0.79}}& {\color{red}\textbf{1.51}}\\
\hline
\end{tabular}
}
\label{tab:fdconfig}
\end{table}

\subsection{Ablation Study}
This section provides a detailed ablation study conducted on the LLVIP dataset for the IVF task to evaluate the effectiveness of the modules proposed in our network.

\begin{table}[t!]
\renewcommand\arraystretch{1.2}
\tabcolsep 2pt
\caption{Ablation Results for AdaD-SSD. We consider the State Space Model (SSM) in Mamba~\cite{mamba} as a benchmark and gradually test state space duality (SSD)~\cite{mamba2}, 2D-SSD, spatial-aware branch $\mathcal{S}(\cdot)$, and frequency filtering branch $\mathcal{T}(\cdot)$.}
\resizebox{1.0\linewidth}{!}{
\begin{tabular}{ll@{\hspace{4pt}}||cccccccc}
\hline
\thickhline
\rowcolor[HTML]{f8f9fa} & Types & EN$\uparrow$ & SD$\uparrow$ & SF$\uparrow$ & MI$\uparrow$ & SCD$\uparrow$ & VIF$\uparrow$ & Qabf$\uparrow$ & SSIM$\uparrow$ \\
\hline
C.I & SSM~\cite{mamba} & 6.88 & 46.35 & 12.59 & 2.00 & 1.38 & 0.69 & 0.61 & 1.38  \\
C.II & SSD~\cite{mamba2} & 7.09 & 45.11 & 13.21 & 2.23 & 1.42 & 0.81 & 0.55 & 1.42  \\
C.III & $\rightarrow$ 2D-SSD &{\color{blue} 7.33} & 47.03 & 14.85 & 2.70 & 1.42 & 0.88 & 0.61 & {\color{blue}1.47} \\
C.IV & $\textit{w}$/ $\mathcal{S}(\cdot)$ & {\color{blue}7.33} & {\color{blue}47.10} & {\color{blue}14.92} & {\color{blue}2.91} & {\color{blue}1.45} & {\color{blue}0.95} & {\color{blue}0.71} & {\color{blue}1.47} \\
\rowcolor{gray!10}C.V & $\textit{w}$/ $\mathcal{T}(\cdot)$\!\!& {\color{red}\textbf{7.39}}& {\color{red}\textbf{47.60}}& {\color{red}\textbf{15.28}}& {\color{red}\textbf{3.06}}& {\color{red}\textbf{1.48}}& {\color{red}\textbf{1.01}}& {\color{red}\textbf{0.79}}& {\color{red}\textbf{1.51}}\\
\hline
\end{tabular}
}
\label{tab:mamba}
\end{table}

\begin{table*}[t!]
\centering 
\renewcommand\arraystretch{1.1}
\caption{Object detection performance of visible (VIS), infrared (IR), and fusion images (our proposed method) on the M3FD dataset. {\color{red}Red} color indicates the best result.}
\resizebox{1.0\textwidth}{!}{  
\begin{tabular}{l||l|ccccccc|ccccccc}
\hline
\thickhline
\rowcolor[HTML]{f8f9fa}
 & & \multicolumn{7}{c|}{YOLOv5} & \multicolumn{7}{c}{YOLOv8} \\
 \rowcolor[HTML]{f8f9fa} & & People & Car & Bus & Motor & Lamp & Truck & Average & People & Car & Bus & Motor & Lamp & Truck & Average \\ \hline
\multirow{3}{*}{{IR}}\!& AP@0.5 &  0.720&     0.835&0.864& 0.531& 0.398& 0.754&     0.671
&  0.757&     0.822&     0.851& 0.496& 0.340& 0.724&     0.650\\
 & AP@0.75     &  {\color{red}\textbf{0.349}} &     0.543&     0.688& 0.257& 0.120&   {\color{red}\textbf{0.565}}&     0.411
&  0.382&     0.562&     0.744& 0.242& 0.136& 0.554&     0.429\\
 & mAP@{[}0.5:0.95{]} &  0.370&     0.520&     0.616&   {\color{red}\textbf{0.302}}& 0.160&  {\color{red}\textbf{0.489}}&  0.403& {\color{red}\textbf{0.403}}&     0.528&     0.645& 0.267& 0.162& 0.488&     0.409\\ \hline

\multirow{3}{*}{{VIS}}\!& AP@0.5 &  0.603&     0.856&     0.886&  {\color{red}\textbf{0.546}}&  {\color{red}\textbf{0.699}}& 0.759&     0.705&  0.625&     0.859&     0.849& 0.514& 0.572& 0.709&     0.677\\
 & AP@0.75     &  0.199&     0.573&     0.687& 0.244& 0.199& 0.519&     0.394&  0.222&     0.621&  {\color{red}\textbf{0.799}}& 0.276& 0.175& 0.471&     0.438\\
 & mAP@{[}0.5:0.95{]} &  0.264&     0.547&     0.646& 0.277& 0.285& 0.470&     0.406&  0.289&     0.572&     0.675& 0.270& 0.253& 0.434&     0.412\\ \hline

\multirow{3}{*}{{Fusion}}\!& AP@0.5 &  {\color{red}\textbf{0.740}}    &   {\color{red}\textbf{0.882}}   &    {\color{red}\textbf{0.908}} & 0.544& 0.663&     {\color{red}\textbf{0.823}}& {\color{red}\textbf{0.746}}&  {\color{red}\textbf{0.774}}& {\color{red}\textbf{0.894}}& {\color{red}\textbf{0.878}}& {\color{red}\textbf{0.612}}&   {\color{red}\textbf{0.706}}&   {\color{red}\textbf{0.829}}& {\color{red}\textbf{0.767}}\\
 & AP@0.75     &  0.343&    {\color{red}\textbf{0.614}} & {\color{red}\textbf{0.746}}&  {\color{red}\textbf{0.290}}& {\color{red}\textbf{0.238}}& 0.545&   {\color{red}\textbf{0.450}}&   {\color{red}\textbf{0.397}}&  {\color{red}\textbf{0.675}}&     0.757&   {\color{red}\textbf{0.317}}&   {\color{red}\textbf{0.259}}&  {\color{red}\textbf{0.620}}&   {\color{red}\textbf{0.487}}\\
 & mAP@{[}0.5:0.95{]} &  {\color{red}\textbf{0.379}} &     {\color{red}\textbf{0.577}}&   {\color{red}\textbf{0.675}}& 0.294&    {\color{red}\textbf{0.292}}& 0.441&   {\color{red}\textbf{0.447}}&  0.319&  {\color{red}\textbf{0.615}}& {\color{red}\textbf{0.693}}&     {\color{red}\textbf{0.327}}&   {\color{red}\textbf{0.325}}&  {\color{red}\textbf{0.559}}&  {\color{red}\textbf{0.480}}\\ \hline
\end{tabular}
}
\label{tab:yolo}
\end{table*}

\begin{table}[t!]
\centering
\renewcommand\arraystretch{1.1}
\tabcolsep 2pt
\caption{Semantic Segmentation performance of visible (VIS), infrared (IR), and fusion images (the proposed method) on the MFNet dataset. {\color{red}Red} color indicates the best result.}
\resizebox{0.5\textwidth}{!}{  
\begin{tabular}{l||c|cccccccc|c}
\hline
\thickhline 
\rowcolor[HTML]{f8f9fa}
&     & Unlabel & Car & Person & Bike & Curve & Car & Cone & Bump & Average \\ \hline
\multirow{2}{*}{IR}  & Acc &  0.99   & 0.90    &   0.79    &   0.60   &  0.34     &     0.24     &  0.38    &  {\color{red}\textbf{0.63}}  &   0.54      \\
    & IoU &   0.98      &  0.84   &   0.67     &   0.52   &   0.29    &   0.19   &  0.27    & {\color{red}\textbf{0.50}}   &   0.47      \\ \hline
\multirow{2}{*}{VIS}   & Acc &   0.99      &   {\color{red}\textbf{0.92}}  &  0.71      &  {\color{red}\textbf{0.76}}    &  0.34     &   {\color{red}\textbf{0.39}}     &  0.51    & 0.55     &   0.57      \\
    & IoU &  0.98   &  0.86   &  0.59  & 0.61     & 0.28     & {\color{red}\textbf{0.31}} &  0.45    &  0.41  &    0.49    \\ \hline
\multirow{2}{*}{Fusion} & Acc & {\color{red}\textbf{0.99}}  &  0.91   &  {\color{red}\textbf{0.80}}     &  0.75    &   {\color{red}\textbf{0.47}}    &   0.27       &  {\color{red}\textbf{0.52}}   & 0.57     & {\color{red}\textbf{0.59}}      \\
    & IoU &  {\color{red}\textbf{0.99}}  & {\color{red}\textbf{0.86}}  &  {\color{red}\textbf{0.69}}   &  {\color{red}\textbf{0.63}}   &  {\color{red}\textbf{0.37}}  &   0.23   &  {\color{red}\textbf{0.48}}   &   0.47 & {\color{red}\textbf{0.52}}    \\ \hline
\end{tabular}
}
\label{tab:Segmentation}    
\end{table}

\subsubsection{Effectiveness of Key Components}
In this ablation study, we use the network that excludes AdaWAT, the Spatial-Frequency Mamba block with shallow fusion $\textbf{M}_{shallow}$, and it as deep fusion $\textbf{M}_{deep}$ as the baseline to assess the contribution of the proposed core modules to the fusion performance.
The results are reported in Table~\ref{tab:module}, where the network with $\textbf{M}_{deep}$ (\ie, the complete AdaSFFuse network) demonstrates superior performance in this task. Specifically, compared to the version without $\textbf{M}_{deep}$, the SSIM score shows a significant improvement, with a value of 1.51 \myvs 1.41.

\subsubsection{Impact of Frequency Decoupling}
To assess the effect of the proposed frequency decoupling module, AdaWAT, we replaced it with commonly used methods, including FFT~\cite{shen2024spatial}, Laplacian decomposition~\cite{iqbal2020underwater}, and wavelet transforms (WAT)~\cite{huang2022winnet}. 
As shown in Table~\ref{tab:fdconfig}, AdaWAT's ability to approximate and enhance frequency features across different modalities leads to comprehensive improvements in the fusion results, validating its more robust and versatile performance.

\subsubsection{Impact of Spatial-Frequency Mamba in Modules}
Here, we evaluated the contributions of the core modules in the Spatial-Frequency Mamba block. The evaluation involved comparing the State Space Model (SSM) in Mamba~\cite{mamba} as the baseline, and progressively testing the impact of the state space duality (SSD)~\cite{mamba2}, 2D-SSD, spatial-aware branch $\mathcal{S}(\cdot)$, and frequency filtering branch $\mathcal{T}(\cdot)$. The results are recorded in Table~\ref{tab:mamba}. The comparison between studies C.II and C.III shows that 2D-SSD, compared to SSD, is more suitable for handling image features, leading to a higher gain in MMIF tasks (\ie, SSIM of 1.47 \myvs 1.42). Furthermore, the studies in C.IV and C.V validate the effective improvement brought by $\mathcal{S}(\cdot)$  and $\mathcal{T}(\cdot)$ in enhancing the perception of local spatial information and the complementary integration of global frequency information.


\subsection{Downstream Task Evaluation}
To evaluate the practical effectiveness of our fusion model, we follow the protocol of~\cite{Liu_2023} and assess its performance on two key vision tasks: object detection and semantic segmentation. These tasks highlight the model’s capacity to produce task-relevant features from multimodal inputs, validating the utility of the fused outputs in real-world applications.

\subsubsection{Performance on Object Detection}
For the object detection task, we utilized 4,200 multimodal images from the M3FD dataset~\cite{liu2022target} to evaluate the fusion quality. YOLOv5~\cite{Jocher_YOLOv5_by_Ultralytics_2020} and YOLOv8~\cite{Jocher_Ultralytics_YOLO_2023} are used as the baseline model for object detection. The experimental results for the visible and infrared images are shown in Table~\ref{tab:yolo}. The results show that our fusion method can effectively facilitate the segmentation model to perceive the imaging scene by fully integrating the complementary intra- and inter-modal information and the global context. In our approach, we can continuously generate detection-friendly fusion results, which is advantageous in detecting people and vehicles, such as obscured cars and pedestrians on distant rocks. In conclusion, the object detection performance experiments validate the value of the fusion results for practical applications.
\subsubsection{Performance on Semantic Segmentation}
For the semantic segmentation task, we conducted experiments on the MFNet~\cite{ha2017mfnet} dataset with the SOTA semantic segmentation model of SegFormer~\cite{xie2021segformer}. 
Following the common practice~\cite{ding2023hgformer,Liu_2023}, we loaded the pre-trained weights \textit{mb1} and fine-tuned them. 
We used pixel Accuracy (Acc) and Intersection-Over-Union (IoU) to evaluate segmentation performance. The experimental results are reported in Table~\ref{tab:Segmentation}.
We can see that the fusion images obtain the highest IoU in the main object categories (\eg, cars and people), ranking first in terms of IoU and Average Acc. 
We attribute this advantage to two factors. On the one hand, our fusion network finely captures valuable frequency information between cross-modal images to promote efficient and synergistic multi-domain fusion. On the other hand, the proposed AdaSFFuse empowers spatial sensing and frequency modulation to efficiently and dynamically adapt semantic feature neutralization to the fusion process, resulting in fused images with rich semantic information.

\begin{figure}[t!]
\centering
\includegraphics[width=1\linewidth]{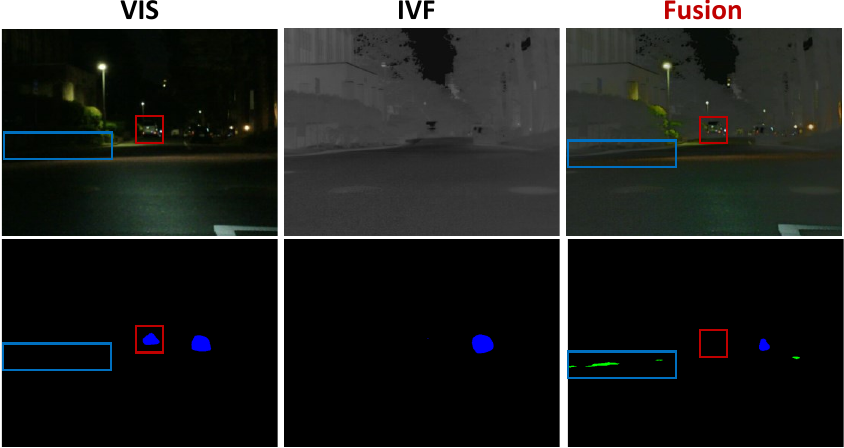}
\caption{A case study illustrating the limitations of fusion-based semantic segmentation on the MFNet dataset. While the proposed AdaSFFuse network introduces richer contextual cues through visible-infrared fusion, certain fine-grained objects (\eg, cars) may become less distinguishable. This highlights the need for more balanced feature integration in future fusion designs.
} 
\label{Fig:case}
\end{figure}


\section{Discussion}
While AdaSFFuse consistently improves fusion quality across diverse scenarios, task-level analysis indicates that enhanced fusion representations do not always translate into improved performance for downstream tasks. In some cases, critical semantic information may be diminished, even when the overall visual clarity of the fused image is improved. This is reflected in Fig.~\ref{Fig:case}, where important target information present in the unimodal input is not preserved after fusion, underscoring the importance of aligning fusion strategies with task-specific objectives.
Rather than treating fusion and task learning as isolated stages, future research may benefit from exploring fully end-to-end, task-aware multimodal frameworks where the fusion process is jointly optimized with respect to task-specific loss functions and structural constraints. Such integration holds promise for building more robust, adaptive models that can dynamically emphasize modality-specific cues depending on scene context and task demands, ultimately leading to more reliable and interpretable multimodal systems.


\section{Conclusion}
In this work, we presented AdaSFFuse, a novel approach for task-generalized multimodal image fusion (MMIF). AdaSFFuse tackled fundamental challenges in multimodal fusion, including high-frequency detail destruction and task-specific limitations. It featured two key innovations: AdaWAT, which enabled fine-grained frequency decoupling, and Spatial-Frequency Mamba blocks, which optimized fusion by leveraging both spatial and frequency domains. AdaSFFuse adapted effectively to diverse modalities and tasks, ensuring robust feature extraction and integration. Extensive experiments on four MMIF tasks—IVF, MFF, MEF, and MIF—demonstrated its superior performance and efficiency, establishing a new benchmark for generalized multimodal fusion.

{
\small
\bibliographystyle{IEEEtran}
\bibliography{IEEEtrans}
}
\begin{IEEEbiography}[{\includegraphics[width=1in,height=1.25in,clip,keepaspectratio]{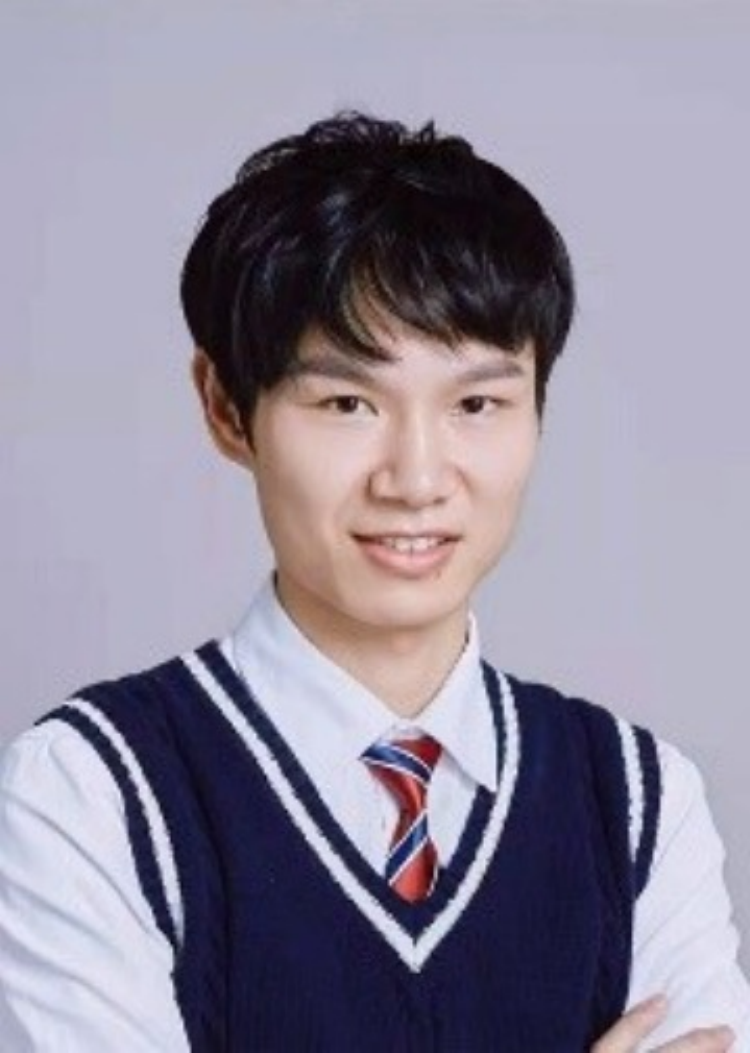}}]{Mengyu Wang} received the B.E. degree in engineering from Harbin Institute of Technology, Harbin, China, in 2015, and the Ph.D degree in engineering from University of Science and Technology of China, Hefei, China, in 2020. He is currently an associate professor with the Key Laboratory of Opto-Electronic Information Science and Technology of Jiangxi Province and the Key Laboratory of Nondestructive Test (Ministry of Education), Nanchang, China. His research interests include computer vision, intelligence sensing, and precise measurement.
\end{IEEEbiography}
\begin{IEEEbiography}[{\includegraphics[width=1in,height=1.25in,clip,keepaspectratio]{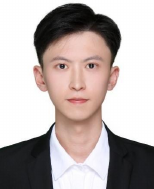}}]{Zhenyu Liu} received the B.E. degree in electronic information from Hunan Engineering College, Xiangtan, China in 2023. He is currently pursuing the M.E. degree in optoelectronic information engineering at Nanchang Hangkong University, Jiangxi, China. His research interests include computer vision and deep learning.
\end{IEEEbiography}
\begin{IEEEbiography}[{\includegraphics[width=1in,height=1.25in,clip,keepaspectratio]{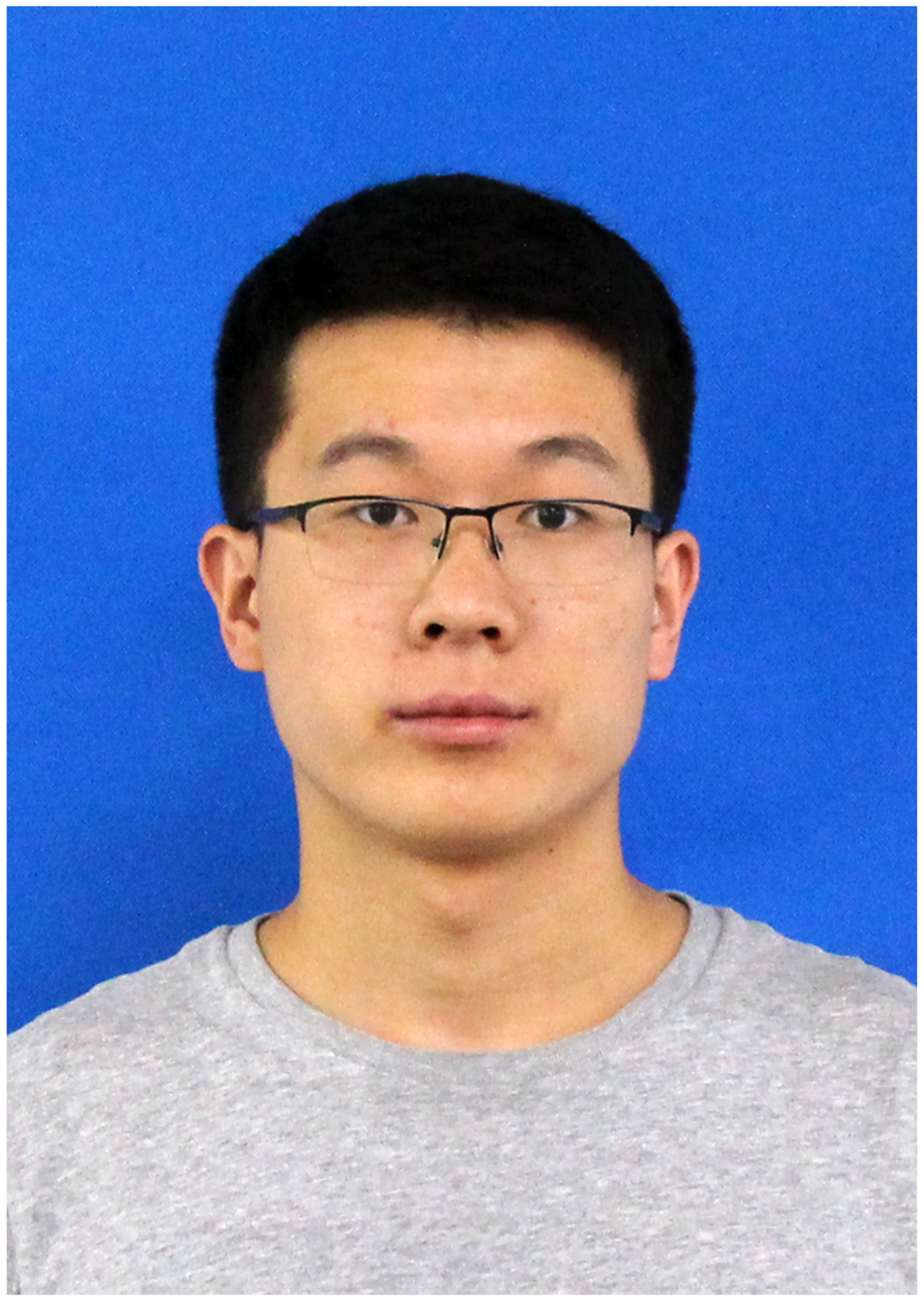}}]{Kun Li} is currently a Research Assistant at Zhejiang University. He received his Ph.D. degree from Hefei University of Technology, in 2023. His research interests include multimedia content analysis, computer vision, and video understanding. He regularly serves as a PC Member for top-tier conferences in multimedia and artificial intelligence, like CVPR, ICCV, ECCV, ACM Multimedia, and AAAI.
\end{IEEEbiography}
\begin{IEEEbiography}[{\includegraphics[width=1in,height=1.25in,clip,keepaspectratio]{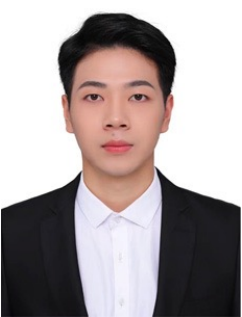}}]{Yu Wang} received the B.E. degree in electrical engineering and automation from Anhui Agricultural University, Hefei, China, in 2022. He is currently pursuing the M.E. degree in agricultural electrification and automation with the College of Engineering, Anhui Agricultural University, Hefei, China. His research interests include computer vision and deep learning.
\end{IEEEbiography}
\begin{IEEEbiography}[{\includegraphics[width=1in,height=1.25in,clip,keepaspectratio]{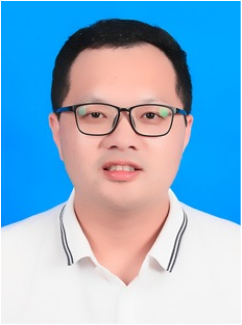}}]{Yuwei Wang} received the B.E. degree in measuring testing technologies and instruments from Hefei University of Technology, Hefei, China, in 2012, and the Ph.D. degree in instrument science and technology from University of Science and Technology of China, Hefei, China, in 2017. He is currently an Associate Professor with the College of Engineering, Anhui Agricultural University, Hefei, China. His research interests include machine vision and 3D shape measurement.
\end{IEEEbiography}
\begin{IEEEbiography}[{\includegraphics[width=1in,height=1.25in,clip,keepaspectratio]{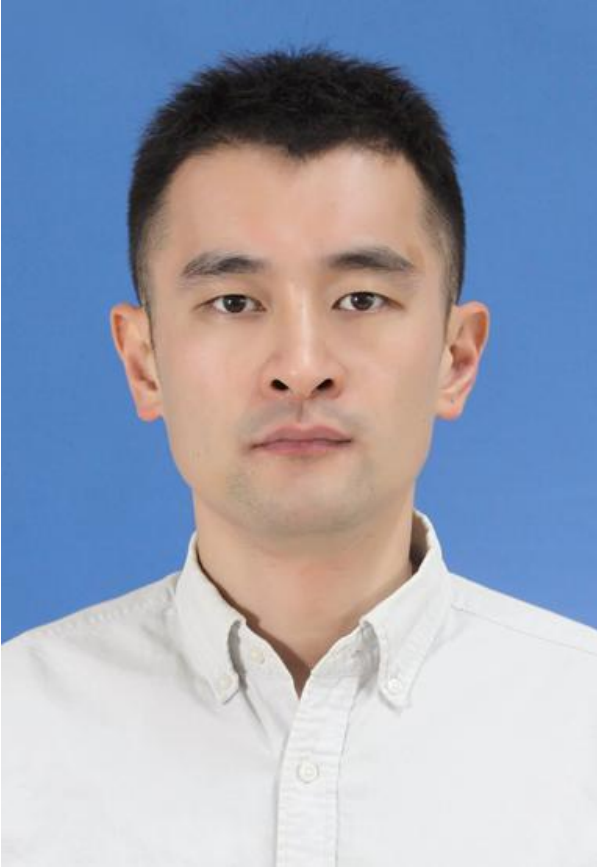}}]{Yanyan Wei} is a Lecturer and Master's Supervisor at the School of Computer Science and Information Engineering, Hefei University of Technology. He earned his Ph.D. in Engineering in December 2022. His research interests include: 1) Image Restoration, Enhancement, and Applications; 2) Image Generation and Style Transfer; 3) Image Understanding and Feature Extraction. He has published 10+ papers in top journals and conferences, including IEEE TIP, IEEE TMM, and ACM MM, and received three competition awards from CVPR and IJCAI.
He is a Program Committee member for AAAI 2024 and a reviewer for leading venues like IJCV, CVPR, ICCV, and ICLR. He is a member of ACM, CCF, CSIG, and CAAI.
\end{IEEEbiography}
\begin{IEEEbiography}[{\includegraphics[width=1in,height=1.25in,clip,keepaspectratio]{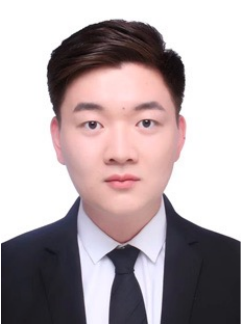}}]{Fei Wang} is currently pursuing the Ph.D. degree in Engineering with the School of Computer Science and Information Engineering, Hefei University of Technology, Hefei, China. His research interests include computer vision and multimodal affective computing. He has published four papers at top international conferences, including CVPR, AAAI, and IJCAI, and received five competition awards from ACM MM and IJCAI. He regularly serves as a PC Member for top-tier conferences in multimedia and artificial intelligence, like ACM MM and ICLR.
\end{IEEEbiography}
\end{document}